\documentclass[letterpaper]{article} 
\usepackage{aaai2027}  
\usepackage[hyphens]{url}  
\usepackage{graphicx} 
\usepackage{natbib}  
\usepackage{caption} 
\usepackage{algorithm}
\usepackage{algorithmic}

\usepackage{newfloat}
\usepackage{listings}
\DeclareCaptionStyle{ruled}{labelfont=normalfont,labelsep=colon,strut=off} 
\floatstyle{ruled}
\newfloat{listing}{tb}{lst}{}
\floatname{listing}{Listing}

\usepackage{booktabs}

\usepackage{amsmath}
\usepackage{amssymb}
\usepackage{mathtools}
\usepackage{amsthm}
\usepackage{multirow}
\usepackage{tcolorbox}
\usepackage{xcolor}
\usepackage{enumitem}
\nocopyright 
		
\title{BrainAgent: A Large Language Model-Driven Multi-Agent Framework \\ for Autonomous Brain Signal Understanding}
\author{
	\thanks{Preprint}
	Yangxuan Zhou\textsuperscript{\rm 1,2}, Sha Zhao\textsuperscript{\rm 1,2\thanks{Corresponding authors}}, Jiquan Wang\textsuperscript{\rm 1},  Shijian Li\textsuperscript{\rm 1,2}, Gang Pan\textsuperscript{\rm 1,2,3}
}
\affiliations{
	\textsuperscript{\rm 1}State Key Laboratory of Brain-machine Intelligence, Zhejiang University, Hangzhou, China\\ \textsuperscript{\rm 2}College of Computer Science and Technology, Zhejiang University, Hangzhou, China\\  \textsuperscript{\rm 3}MOE Frontier Science Center for Brain Science and Brain-machine Integration, Zhejiang University, Hangzhou, China \\ 
	\{zyangxuan, szhao, wangjiquan, shijianli, gpan\}@zju.edu.cn
}

\begin{document}

\maketitle

\begin{abstract}
	Brain-Computer Interfaces (BCIs) and brain signal understanding are pivotal for clinical health and next-generation interactions. Despite this significance, its widespread adoption in real-world scenarios remains restricted, primarily because current analytical paradigms lack sufficient agentic intelligence. First, existing methodologies impose prohibitive technical barriers, requiring extensive specialized expertise. Second, they remain inherently static and task-specific, failing to execute the complex, long-horizon workflows essential for real-world deployment. To accelerate the democratization of brain signal understanding, we draw inspiration from Large Language Models (LLMs) to introduce BrainAgent, an LLM-driven multi-agent framework designed to ground abstract natural language intent into rigorous, executable, and end-to-end processing pipelines. BrainAgent employs a hierarchical architecture where a central supervisor orchestrates specialized sub-agents for adaptive task decomposition and execution. Furthermore, we establish a comprehensive, systematic benchmark for evaluating agentic systems in brain signal analysis. Empirical results demonstrate that BrainAgent effectively automates complex workflows with superior reliability, marking a paradigm shift toward democratized brain signal understanding.
\end{abstract}


\section{Introduction}
Brain-Computer Interfaces (BCIs) serve as a vital conduit bridging the human brain and the external world, decoding underlying neural mechanisms to enable direct communication and control \cite{schalk2004bci2000, wu2016cyborg, kapitonova2022framework}. Central to this domain are brain signals, which encapsulate rich physiological information reflecting brain activities \cite{waldert2009review}. Among various modalities, Electroencephalography (EEG) has emerged as the predominant choice due to its non-invasiveness and high temporal resolution. In practice, EEG analysis underpins a wide range of critical applications, including clinical sleep staging \cite{zhou2025personalized}, fatigue monitoring \cite{charbonnier2016eeg, hu2017comparison}, cognitive workload assessment \cite{zhou2021cognitive, chikhi2022eeg}, and emotion recognition \cite{li2022eeg, chen2025emod}. Historically, the paradigm of brain signal understanding has evolved significantly. Initially, the field was dominated by a feature-engineering paradigm, where approaches relied heavily on hand-crafted features combined with traditional classifiers \cite{lotte2007review}, necessitating extensive domain expertise to decode brain activities. The advent of deep learning marked a shift towards a representation-decoding paradigm \cite{lecun2015deep}, where end-to-end models achieved superior decoding performance by automatically extracting latent features from EEG signals \cite{lawhern2018eegnet, song2022eeg, wang2024cbramod}.

Despite these evolutionary strides, the current decoding-centric paradigm of brain signal understanding lacks sufficient agentic intelligence, which has significantly hindered its widespread adoption in real-world scenarios. Specifically, existing BCI analytical framework remain constrained by two fundamental bottlenecks. \textbf{First, prohibitive technical barriers impede the democratization of brain signal understanding.} Current brain signal analysis demand extensive specialized expertise (e.g., programming proficiency, signal processing knowledge), requirements that often limit the participation of clinicians and general researchers. These disconnects severely obstruct the path toward generalized and democratized brain signal understanding. \textbf{Second, current analytical paradigms are inherently static and task-specific.} Practical brain signal understanding mandates complex, long-horizon workflows rather than isolated decoding events. A clear example is clinical sleep diagnosis, which involves a rigorous sequential pipeline: from data loading and preprocessing to sleep staging and report generation. However, existing decoding-centric paradigms lack the comprehensive capabilities to execute such continuous, end-to-end brain signal processing pipelines. \textbf{Addressing these limitations necessitates a paradigm shift from isolated solvers to an agentic framework capable of autonomous reasoning and comprehensive execution.} Fortunately, emerging Large Language Models (LLMs) \cite{brown2020language, ouyang2022training} offer a transformative pathway. Distinguished by their advanced planning and tool utilization, LLMs can function as intelligent agents to autonomously orchestrate complex analytical workflows, effectively translating human intent into rigorous brain signal processing pipelines essential for real-world applications.

Building on this potential, recent perspectives have envisioned extending BCI research beyond simple command decoding, positioning LLM-powered agents as active partners capable of interpreting complex cognitive states \cite{kapitonova2024human, chen2025embracing, lee2025brain, baradari2025neurochat, jin2025innovative, li2025multimodal}. Pioneering efforts such as EEGAgent \cite{zhao2025eegagent} have realized this vision with an agent-based framework that enables basic event detection and automated reporting. Despite significant progress, developing a robust agentic framework for brain signal understanding still faces two critical challenges: \textbf{1) the dynamic translation of natural language intent into executable pipelines for democratized brain signal understanding}, and \textbf{2) the establishment of systematic benchmarks to rigorously validate the performance and reliability of such agentic systems.}

To bridge these gaps, we introduce \textbf{BrainAgent, an LLM-driven multi-agent framework} designed to democratize brain signal analysis by grounding natural language instructions into comprehensive, end-to-end processing pipelines essential for real-world scenarios. BrainAgent adopts a hierarchical architecture, where a central supervisor orchestrates specialized sub-agents capable of collaborative reasoning and logical interdependence. This design allows for the adaptive decomposition of complex, long-horizon workflows into manageable sub-tasks. Such modularity ensures robust extensibility across evolving analytical needs and various brain signal modalities. Furthermore, \textbf{we systematically establish a comprehensive benchmark} for evaluating our agentic systems. Our contributions are as follows:

\begin{itemize}
	\item 
	We propose \textbf{BrainAgent, an LLM-driven multi-agent framework that grounds natural language intents into rigorous brain processing pipelines.} By automating end-to-end workflows, it effectively democratizes access to advanced brain signal analysis, overcoming the rigidity and high technical barriers of prior analytical paradigms.
	
	\item 
	We establish a \textbf{systematic, hierarchical benchmark specifically designed to evaluate agentic systems in brain signal analysis.} Spanning from atomic instruction execution to complex long-horizon reasoning, this benchmark serves as a rigorous standard to validate agentic systems in realistic deployment settings.
	
	\item 
	We devise a \textbf{hierarchical architecture wherein a central supervisor orchestrates specialized sub-agents.} This design enables the adaptive decomposition of complex execution workflows and ensures robust extensibility, allowing the framework to seamlessly incorporate new capabilities as analytical demands evolve across diverse brain signal domains.
\end{itemize}

\section{Related Work}
\subsection{Brain Signal Analysis Paradigms}
Standard brain signal analysis, particularly for EEG, typically follows a sequential pipeline comprising data acquisition, preprocessing, feature extraction, and downstream tasks such as classification or diagnosis. Early methodologies relied heavily on domain expertise to construct hand-crafted features \cite{mcfarland2006bci, lotte2007review, bashashati2007survey}, which often suffer from poor generalization. The emergence of deep learning enables end-to-end models to automatically learn latent representations from raw signals and achieving superior performance in diverse brain analysis tasks \cite{supratak2017deepsleepnet, wang2018lstm, peh2022transformer}. However, these approaches are inherently task-specific, lacking the flexibility to handle open-ended, complex instructions beyond their trained scope. Furthermore, the deployment of such workflows typically depends on comprehensive toolboxes like EEGLAB \cite{delorme2004eeglab} and MNE \cite{gramfort2014mne}. While these toolboxes offer robust functionality, they necessitate substantial programming proficiency and signal processing knowledge. This high technical barrier creates a disconnect between high-level human intent and low-level pipeline execution, underscoring the need for more accessible and automated analysis frameworks.
\begin{figure*}[!ht]
	\centering
	\includegraphics[width=1.0\textwidth]{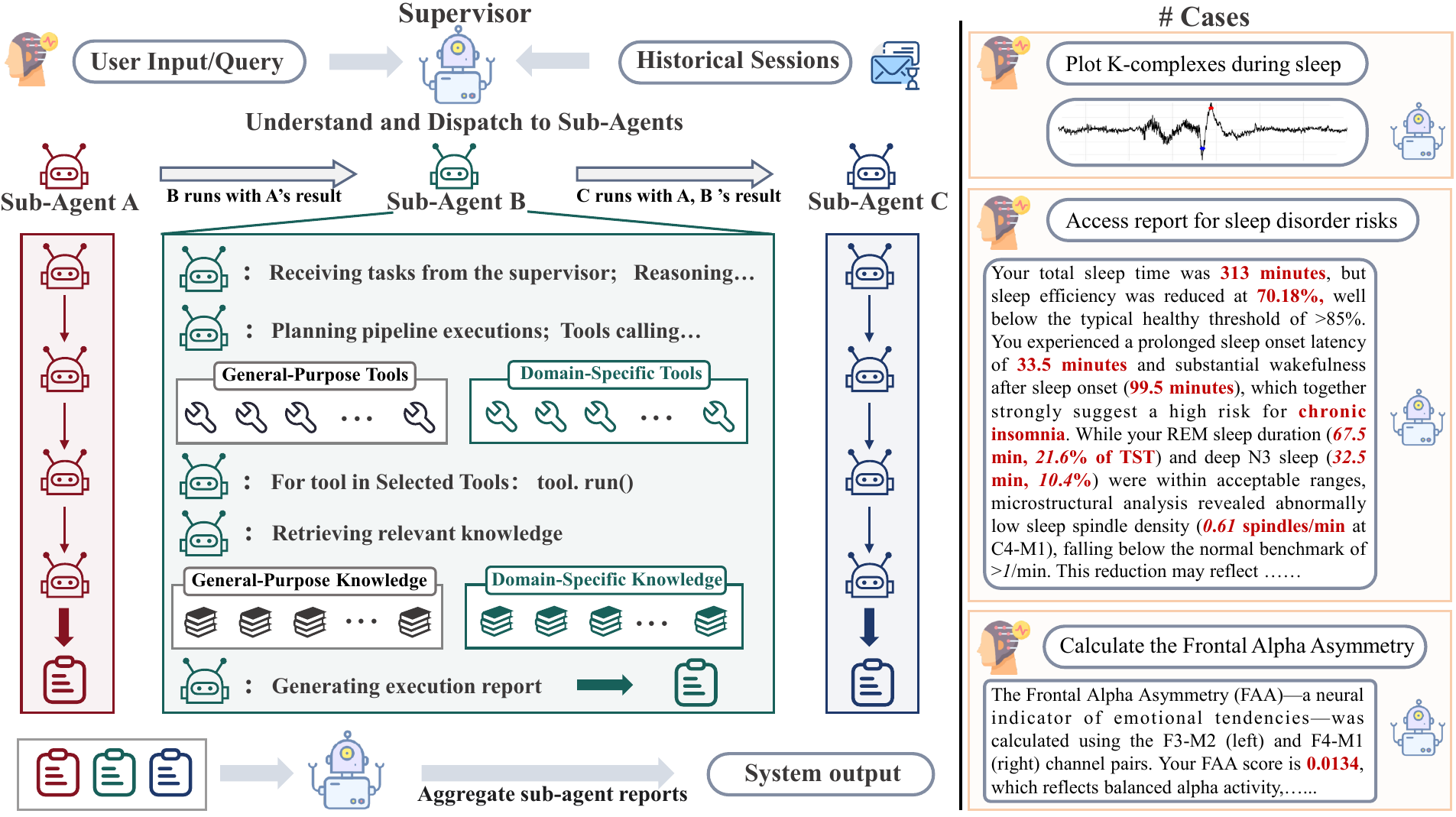}
	\caption{Overview of the BrainAgent framework: The supervisor interprets user queries and dispatches modular sub-agents to execute domain-specific tasks in sequence, leveraging general-purpose and domain-specific tools and knowledge bases. Sub-agent outputs are aggregated into a synthesized system response. Illustrative cases are shown on the right panel. Note that the output report is dynamically generated by BrainAgent based on tool execution analysis, rather than being a templated output.}
	\label{fig:overview}
\end{figure*}
\subsection{Large Language Models for BCI}
Recent research has explored integrating LLMs to enhance BCIs, primarily focusing on cross-modal translation such as decoding EEG signals into text \cite{wang2022open, duan2023dewave, mishra2025thought2text}, image \cite{singh2023eeg2image, ferrante2024decoding} or videos \cite{liu2024eeg2video, huang2025need}. While these approaches effectively leverage the generative power of LLMs, they typically treat the model as a passive decoder rather than an autonomous agent capable of decision-making. Recognizing this limitation, \citet{chen2025embracing} have advocated for a paradigm shift towards Brain-Agent Collaboration, positioning agents as active partners for intelligent assistance rather than mere signal processors; NeuroChat \cite{baradari2025neurochat} enables neuroadaptive tutoring through closed-loop engagement tracking; ChatBCI \cite{kapitonova2024human} integrates LLMs to facilitate human-AI collaboration in BCI research; and EEGAgent \cite{zhao2025eegagent} introduces an agentic framework for EEG event detection and report generation. However, despite representing significant progress, these pioneering efforts generally operate within predefined, relatively rigid workflows and often lack systematic benchmarks to rigorously evaluate their performance in complex, open-ended analytical scenarios. This gap underscores the need for more flexible, autonomous architectures capable of handling the complexities of real-world brain signal analysis.

\subsection{LLM-Driven Autonomous Agents}
The rapid advancement of LLMs has catalyzed the development of LLM-driven autonomous agents, extending their utility from text generation to the domain of autonomous action \cite{yao2022react, wei2022chain}. These systems are characterized by core competencies including autonomous planning \cite{shinn2023reflexion, huang2024understanding}, tool utilization \cite{schick2023toolformer, m2024augmenting}, and the ability to perceive and interact with external environments \cite{yao2022webshop, wu2024copilot}. To handle increasingly complex tasks, research has shifted towards multi-agent collaboration \cite{chan2023chateval, hong2023metagpt, zhang2023building, guo2024large, wu2024autogen}, where specialized agents operate within hierarchical or cooperative structures to enhance stability and problem-solving efficiency compared to single-agent baselines. Consequently, LLM-based agents have been widely deployed across diverse fields, such as chemistry \cite{bran2023chemcrow, yu2024chemtoolagent}, biology \cite{xiao2024cellagent, jin2025stella}, and healthcare \cite{qiu2024llm, wang2025survey}, etc. Drawing inspiration from these successes, we propose BrainAgent, an LLM-driven multi-agent framework that enables democratized, end-to-end autonomous brain signal understanding.

\section{Methodology}
\subsection{Overview}
Given the vast diversity of signal modalities and downstream applications, constructing an all-encompassing system from the outset is inherently impractical. Consequently, an intuitive solution is to prioritize an extensible architecture capable of adapting to evolving analytical needs. \textbf{To this end, we propose BrainAgent, a hierarchical and highly extensible multi-agent framework designed to automate end-to-end brain signal analysis}, where each sub-agent specializes in a distinct domain. As illustrated in Fig. \ref{fig:overview}, the system operates through structured collaboration between a central supervisor agent $\mathcal{A}_{sup}$ and specialized sub-agents $\mathcal{A}_{sub} = \{\mathcal{A}_1, \dots, \mathcal{A}_k\}$.
Formally, given a natural language query $Q$ and the corresponding brain signal file $D$, the objective is to generate a comprehensive analysis result $R$. To facilitate consistent data flow across the hierarchy, we maintain a global variable $\mathcal{V}_{shared}$ (shared state) to manage the input parameters and intermediate artifacts for all tool executions.
The supervisor $\mathcal{A}_{sup}$ first evaluates $Q$ to determine if it requires domain-specific analysis. If so, it decomposes the query and routes sub-tasks $\{\mathcal{T}_1, \dots, \mathcal{T}_k\}$ to a selected set of sub-agents $\{\mathcal{A}_1, \dots, \mathcal{A}_k\}$, yielding the assignment set $\mathcal{S} = \{(\mathcal{A}_1, \mathcal{T}_1), \dots, (\mathcal{A}_k, \mathcal{T}_k)\}$. Each routed agent $\mathcal{A}_i$ ($1 \leq i \leq k$) engages in an iterative reasoning process, planning and executing a sequence of tools $\{t_{i,1}, \dots, t_{i,m}\}$ from its own toolset to fulfill the corresponding sub-task $\mathcal{T}_i$. 
Upon completion, each sub-agent synthesizes tool outputs and retrieved domain knowledge into a sub-report $r_i$. 
Finally, the supervisor aggregates these sub-reports $\{r_1, \dots, r_k\}$ to construct the cohesive final response $R$, effectively grounding the high-level user intent into rigorous analytical outcomes. 

\subsection{Supervisor: Orchestration and Intent Reasoning}
\label{sec:supervisor}
The supervisor agent $\mathcal{A}_{sup}$ functions as the framework's central orchestrator, \textbf{prioritizing high-level intent reasoning over low-level execution}. Possessing a global view of the conversation history and context, $\mathcal{A}_{sup}$ first evaluates whether the query $Q$ can be resolved using existing state information; if sufficient context is available, it generates a direct response to minimize computational overhead. Conversely, when external execution is required, $\mathcal{A}_{sup}$ operates under a strict \textbf{tool-free policy}. Devoid of analytical tools and knowledge bases, it relies exclusively on linguistic reasoning to decompose $Q$ into a structured plan and delegate sub-tasks to specialized agents. This constraint enforces a rigorous separation of concerns, ensuring the supervisor maintains global alignment with user intent while abstracting away the complexities of execution.
Upon the completion of sub-tasks, the supervisor transitions from orchestrator to synthesizer. It aggregates the heterogeneous intermediate reports $\{r_1, \dots, r_k\}$ returned by the sub-agents, performing cross-domain reasoning to identify intrinsic correlations. Rather than simply concatenating outputs, the supervisor interprets the causal relationships between distinct sub-reports—for instance, linking elevated anxiety markers identified by the \texttt{EmotionAgent} with sleep fragmentation detected by the \texttt{SleepAgent} to diagnose anxiety-induced insomnia. It ensures the final response $R$ provides a holistic and medically coherent analysis. 

\subsection{Sub-Agents: Specialization and Task Executing}
\label{sec:subagents}

In contrast to the supervisor, sub-agents $\mathcal{A}_{sub}$ function as specialized executors designed for strict instruction adherence, \textbf{focusing on low-level decision-making and pipeline execution}. To support this role, we design the sub-agents around two core mechanisms: \textbf{tiered resource access} and \textbf{context isolation}. First, each sub-agent operates within a structured access system that integrates both shared and specialized capabilities (as shown in Figure~\ref{fig:overview}). Specifically, agents are granted access to a multi-layered repository of tools and knowledge bases. For instance, the \texttt{SleepAgent} can utilize shared foundational tools (e.g., \texttt{EEGFileLoader}, \texttt{EEGPreprocessor}) alongside domain-exclusive tools (e.g., \texttt{SleepStager}), while referencing both general EEG concepts (e.g., frequency band) and domain-specific protocols (e.g., AASM manuals \cite{berry2017aasm}). Complementing this resource structure is a strict context isolation protocol enforced during runtime. Unlike the supervisor, sub-agents are blinded to the global conversation history, receiving only the specific task directive and relevant execution outputs from preceding agents. This isolation prevents scope creep, ensuring the sub-agent operates as a focused expert unburdened by the ambiguity of high-level dialogue. 

Collectively, this design yields three critical advantages: first, it enables the supervisor to route tasks to the most appropriate domain experts for rigorous analysis; second, it significantly reduces the context length consumed by extensive tool descriptions, thereby lowering computational costs and mitigating hallucinations; and third, it facilitates exceptional extensibility, allowing the community to seamlessly integrate new tools or agents for diverse BCI applications. 
The algorithmic pipeline is illustrated in Algorithm \ref{alg:brainagent}.
\begin{algorithm}[!t]
	\caption{BrainAgent Execution Workflow}
	\label{alg:brainagent}
	\begin{algorithmic}[1]
		\REQUIRE {User Query $Q$, Brain Signal Files $D$}
		\ENSURE {Final Response $R$}
		\\
		\STATE \textbf{Stage 1: Supervisor Reasoning and Dispatching}
		\STATE $\mathcal{A}_{sup}$ decomposes $Q$ and dispatches sub-tasks: $\mathcal{S} \leftarrow \{(\mathcal{A}_1, \mathcal{T}_1), \dots, (\mathcal{A}_k, \mathcal{T}_k)\}$.
		\\
		\STATE \textbf{Stage 2: Sub-Agent Execution Loop}
		\FORALL{$(\mathcal{A}_i, \mathcal{T}_i)$ in $\mathcal{S}$}
		\STATE Sub-agent $\mathcal{A}_{i}$ reasons and plans the tool execution order $\textrm{T}_i \leftarrow \{t_{i,1}, \dots, t_{i,m}\}$.
		\FORALL{$t_{i,j}$ in $\textrm{T}_i$}
		\STATE Execute tools $t_{i,j}$ on $D$ and update execution trace.
		\ENDFOR
		\STATE  Sub-agent $\mathcal{A}_{i}$ generate sub-report $r_i$.
		\ENDFOR
		\\
		\STATE \textbf{Stage 3: Synthesis and Response}
		\STATE The supervisor $\mathcal{A}_{sup}$ aggregates sub-reports $\{r_1, \dots, r_k\}$ to generate final response $R$.
		\STATE \textbf{return} $R$
	\end{algorithmic}
\end{algorithm}
\subsection{Hierarchical Evaluation Benchmark}
Despite the growing interest in agentic systems for brain analysis, there remains a \textbf{notable absence of systematic benchmarks} capable of evaluating their reliability and performance in complex, open-ended scenarios. To address this limitation, we establish a comprehensive, hierarchical benchmark designed to rigorously assess BrainAgent across three critical dimensions: task completion rate, routing accuracy, and tool call accuracy. In this study, we develop the framework with two representative sub-agents: the \texttt{SleepAgent}, encompassing functionalities such as sleep staging, micro-event detection, clinical report generation, etc.; and the \texttt{EmotionAgent}, covering emotional analysis, mental fatigue monitoring, cognitive workload assessment, etc. Grounded in the operational boundaries of these agents and their respective toolsets, we categorize analytical tasks into three difficulty levels based on the ambiguity of the instruction and the complexity of the execution pipelines.

\subsubsection{Task Complexity Levels} 

To systematically evaluate the capability boundaries of the BrainAgent framework, we stratify analytical tasks into three distinct levels of complexity. This stratification is essential to disentangle different aspects of agentic intelligence: \textbf{Level 1} isolates the fundamental ability to execute explicit commands and correct tool parameterization; \textbf{Level 2} evaluates the capacity for logical reasoning and long-horizon planning within a single domain; and \textbf{Level 3} tests the system's high-level human intent understanding and cross-domain orchestration. This hierarchical approach allows for a granular diagnosis of system performance, distinguishing between simple execution failures and complex reasoning deficits.

\begin{itemize}
	\item \textbf{Level 1: Atomic Capability (L1).} This level evaluates the agent's proficiency in executing explicit, few-step instructions using basic tools. Tasks typically involve data loading, preprocessing, or other straightforward operations (e.g., \textit{“Load the data and visualize the first 30 seconds of EEG signals”}). It focuses on tool parameter correctness and execution stability within a single sub-agent.
	
	\item \textbf{Level 2: Sequential Reasoning (L2).} This level assesses the agent's ability to plan and execute long-horizon workflows within a single domain, testing its capacity to accurately interpret and execute complex, multi-step instructions. These tasks imply strict logical dependencies where the output of one tool serves as the prerequisite for another (e.g., \textit{“Perform sleep staging, detect spindles in the N2 stage, and generate a summary report”}). Success requires the agent to adhere to a precise operational sequence, ensuring no critical tools are omitted while avoiding the hallucination of non-existent dependencies.
	
	\item \textbf{Level 3: Intent Reasoning and Collaborative Intelligence (L3).} This level represents the highest complexity, simulating real-world scenarios driven by ambiguous, high-level user intents rather than explicit instructions. Tasks often involve abstract queries (e.g., \textit{“Assess the patient's risk for anxiety-induced insomnia”}) that necessitate complex, long-horizon workflows and, in many cases, cross-domain collaboration. Success at this level depends on the system's comprehensive ability to decompose abstract goals, accurately coordinate multiple agents, and synthesize outputs into a coherent analysis.
\end{itemize}

\begin{table*}[!ht]
	\centering
	\caption{Main results of BrainAgent across different backbones and difficulty levels (L1--L3). \textbf{TCR}: Task Completion Rate; \textbf{R-ACC}: Routing Accuracy; \textbf{TCE}: Tool Call Eccuracy; \textbf{Lat.}: Average time cost per successful case (min). The best performance in TCR columns is marked in \textbf{bold}, and the second-best is \underline{underlined}.}
	\label{tab:main_results}
	\resizebox{\textwidth}{!}{%
		\begin{tabular}{l c cccc cccc cccc c}
			\toprule
			\multirow{2}{*}{\textbf{Model}} & \multirow{2}{*}{\textbf{Open Source}} & \multicolumn{4}{c}{\textbf{L1}} & \multicolumn{4}{c}{\textbf{L2}} & \multicolumn{4}{c}{\textbf{L3}} & \textbf{Average} \\
			\cmidrule(lr){3-6} \cmidrule(lr){7-10} \cmidrule(lr){11-14} \cmidrule(lr){15-15}
			& & \textbf{TCR} & \textbf{R-ACC} & \textbf{TCE} & \textbf{Lat.} & \textbf{TCR} & \textbf{R-ACC} & \textbf{TCE} & \textbf{Lat.} & \textbf{TCR} & \textbf{R-ACC} & \textbf{TCE} & \textbf{Lat.} & \textbf{TCR} \\
			\midrule
			Qwen-Flash & $\times$ & 0.75±0.06 & 0.91±0.03 & 0.93±0.01 & 0.59  & 0.77±0.04 & 0.92±0.03 & 0.94±0.01 & 0.94 & 0.58±0.05 & 0.76±0.03 & 0.72±0.02 & 1.57 & 0.70 \\
			Qwen-Plus & $\times$ & 0.83±0.05 & 0.92±0.04 & 0.95±0.02 & 0.92 & 0.91±0.03 & 0.94±0.02 & 0.98±0.01 & 1.22 & 0.67±0.02 & 0.87±0.02 & 0.83±0.03 & 2.10 & 0.80 \\
			Qwen-Max & $\times$ & \textbf{0.95±0.01} & 0.99±0.01 & 0.96±0.01 & 1.05 & \textbf{0.97±0.03} & 1.00±0.00 & 0.98±0.01 & 1.30 & \underline{0.77±0.04} & 0.97±0.02 & 0.87±0.02 & 2.30 & \textbf{0.90} \\
			GPT-4o & $\times$ & 0.87±0.03 & 0.99±0.01 & 0.94±0.02 & 0.60 & 0.89±0.01 & 0.91±0.03 & 0.99±0.01 & 0.94 & 0.68±0.02 & 0.98±0.01 & 0.93±0.02 & 1.24 & 0.81 \\
			\midrule
			Qwen3-8B & $\checkmark$ & 0.56±0.04 & 0.89±0.08 & 0.97±0.01 & 0.62 & 0.47±0.05 & 0.71±0.02 & 0.97±0.01 & 0.76 & 0.34±0.02 & 0.83±0.04 & 0.87±0.02 & 1.19 & 0.46 \\
			Qwen3-30B-A3B & $\checkmark$ & 0.62±0.06 & 0.87±0.05 & 0.98±0.01 & 0.66 & 0.69±0.04 & 0.73±0.03 & 0.99±0.01 & 0.86 & 0.51±0.08 & 0.94±0.03 & 0.90±0.02 & 1.49 & 0.61 \\
			Qwen3.5-122B-A10B & $\checkmark$ & 0.92±0.02 & 0.99±0.01 & 0.97±0.01 & 0.81 & \underline{0.96±0.02} & 0.98±0.02 & 0.99±0.01 & 1.07 & 0.72±0.03 & 0.88±0.02 & 0.89±0.01 & 1.83 & \underline{0.87} \\
			Llama 3.3 70B & $\checkmark$ & 0.85±0.03 & 0.98±0.01 & 0.91±0.02 & 0.75 & 0.60±0.04 & 0.80±0.02 & 0.97±0.01 & 1.22 & 0.40±0.03 & 0.85±0.02 & 0.92±0.03 & 1.68 & 0.62 \\
			DeepSeek-v3.2 & $\checkmark$ & \underline{0.94±0.01} & 0.99±0.01 & 0.94±0.02 & 1.81 & 0.84±0.02 & 0.87±0.03 & 0.95±0.01 & 2.53 & \textbf{0.79±0.01} & 0.91±0.02 & 0.85±0.02 & 4.06 & 0.86 \\
			\bottomrule
		\end{tabular}%
	}
\end{table*}
\subsubsection{Evaluation Metrics} 

To quantitatively assess the framework, we define three core metrics: Task Completion Rate (TCR), Routing Accuracy (R-ACC), and Tool Call Efficiency (TCE). \textbf{TCR} measures the percentage of tasks successfully executed from end to end without runtime errors, defined as:
\begin{equation}
	\text{TCR} = \frac{C}{N} \times 100\%,
\end{equation}
where $C$ is the number of successfully completed tasks and $N$ is the total number of tasks. \textbf{R-ACC} evaluates the supervisor's intent interpretation by calculating the ratio of correct agent dispatch decisions:
\begin{equation}
	\text{R-ACC} = \frac{R}{N} \times 100\%,
\end{equation}
where $R$ denotes the count of tasks where the dispatched agent set matches the ground truth. \textbf{TCE} assesses execution efficiency by measuring the proportion of necessary tools within the execution trace. TCE is averaged across the $C$ successfully completed tasks:
\begin{equation}
	\text{TCE} = \frac{1}{C} \sum_{i=1}^{C} \left( \frac{|\textrm{T}_{correct}^{(i)}|}{|\textrm{T}_{total}^{(i)}|} \right) \times 100\%,
\end{equation}
where $|\mathcal{T}_{correct}^{(i)}|$ and $|\mathcal{T}_{total}^{(i)}|$ represent the count of necessary and total tools executed for task $i$, respectively. It is important to note that owing to disparities in evaluation benchmarks and the definitions of tool capability boundaries, direct comparisons with existing methods (e.g., EEGAgent) are not feasible on our proposed benchmark. Detailed tool descriptions, knowledge base descriptions, benchmark dataset examples, agent system prompts, error analysis and case studies, and output quality evaluation \textbf{are provided in the Appendix.}

\section{Experiments}
\subsection{Experimental Setup}
\textbf{Dataset and Evaluation Protocol.} We evaluate BrainAgent on two classical publicly available sleep datasets, ISRUC Subgroup-1 \cite{khalighi2016isruc} and HMC \cite{hmc}, across varying task difficulties (L1--L3). To improve the robustness of the evaluation and reduce potential subject-specific bias, we select six subjects from ISRUC and four subjects from HMC, resulting in ten subjects in total for repeated experiments. The evaluation benchmark consists of 60 distinct tasks, with 20 tasks for each difficulty level. For each subject-task pair, we conduct 5 independent runs to mitigate the inherent stochasticity of LLM generation, yielding a total of 3,000 execution traces for statistical analysis.

\textbf{Model Baselines.} To comprehensively assess the impact of model capability and scaling on agent performance, we benchmark BrainAgent across a representative selection of LLMs. Our evaluation spans both open-source models (Qwen3-8B, Qwen3-30B-A3B, Qwen3.5-122B-A10B, Llama3.3-70B Instruct, DeepSeek-v3.2) and closed-source proprietary models (Qwen-Flash, Qwen-Plus, Qwen3-Max, GPT-4o) \cite{hurst2024gpt, deepseek, grattafiori2024llama, yang2025qwen3, qwen35blog}.

\subsection{Results}
\subsubsection{Overview Performance}

Table \ref{tab:main_results} presents the quantitative evaluation of BrainAgent across varying difficulty levels. Our analysis reveals three primary findings regarding the capabilities of LLMs within our autonomous brain signal analysis framework. \textbf{First, Qwen-Max sets the current performance upper bound}, achieving the highest average TCR of 0.90, largely due to its near-perfect execution on atomic (L1) and sequential (L2) tasks. Notably, DeepSeek-v3.2 demonstrates superior reasoning in high-complexity scenarios, surpassing all proprietary models with a TCR of 0.79 in hybrid L3 tasks. These results validate the framework's design, proving that BrainAgent can reliably translate abstract human intents into precise execution pipelines when supported by a capable reasoning backbone. \textbf{Second, the effective deployment of BrainAgent is critically dependent on model capacity.} We observe monotonic improvements across the Qwen series and substantial gains between the open-source 8B and 30B variants. This indicates that the sophisticated workflow orchestration required by BrainAgent is non-trivial; the framework necessitates models with sufficient parameter scale to fully unlock its potential for autonomous multi-step analysis.\textbf{Third, the performance disparity between small and large models widens as task complexity increases.} While lightweight models like Qwen3-8B maintain moderate functionality in simple L1 tasks, their completion rates collapse to 0.34 under the ambiguity and dependency constraints of L3. In contrast, frontier models retain high stability. This divergence suggests that while basic instruction following is accessible to smaller models, the complex planning required for clinical analysis remains the exclusive capability of large-scale foundation models. We also report the average latency for successful executions. Generally, the time cost scales proportionally with task complexity across most models. Notably, DeepSeek-v3.2 exhibits a significantly higher temporal cost compared to other backbones. This disparity necessitates a critical re-evaluation of the trade-off between its reasoning performance and the associated latency.

\subsubsection{Efficacy of Heterogeneous Architectures}
\begin{table}[!t]
	\centering
	\caption{ Impact of Supervisor capability on Sub-agent performance. We compare Task Completion Rates (TCR) when sub-agents are orchestrated by themselves (Homogeneous) versus by the stronger Qwen-Max (Heterogeneous).}
	\label{tab:heterogeneous_results}
	\resizebox{1.0\columnwidth}{!}{%
		\begin{tabular}{l l c c c c}
			\toprule
			\multirow{2}{*}{\textbf{Sub-agent}} & \multirow{2}{*}{\textbf{Supervisor}} & \multicolumn{4}{c}{\textbf{Task Completion Rate (TCR)}} \\
			\cmidrule(lr){3-6}
			& & \textbf{L1} & \textbf{L2} & \textbf{L3} &  \multirow{1}{*}{\textbf{Average}} \\
			\midrule
			\multirow{2}{*}{Llama 3.3 70B} & Llama 3.3 70B & 0.85±0.03 & 0.60±0.04 & 0.40±0.03 & 0.62 \\
			& Qwen-Max & \textbf{0.86±0.02} & \textbf{0.82±0.03} & \textbf{0.46±0.04} & \textbf{0.71} \\
			\midrule
			\multirow{2}{*}{Qwen3-8B} & Qwen3-8B & 0.56±0.04 & 0.47±0.05 & 0.34±0.02 & 0.46 \\
			& Qwen-Max & \textbf{0.74±0.02} & \textbf{0.89±0.03} & \textbf{0.48±0.02} & \textbf{0.70} \\
			\midrule
			\multirow{2}{*}{Qwen3-30B-A3B} & Qwen3-30B-A3B & 0.62±0.06 & 0.69±0.04 & 0.51±0.08 & 0.61 \\
			& Qwen-Max & \textbf{0.77±0.03} & \textbf{0.95±0.04} & \textbf{0.68±0.02} & \textbf{0.80} \\
			\midrule
			\multirow{2}{*}{Qwen-Flash} & Qwen-Flash & 0.75±0.06 & 0.77±0.04 & 0.58±0.05 & 0.70 \\
			& Qwen-Max & \textbf{0.87±0.01} & \textbf{0.98±0.02} & \textbf{0.67±0.01} & \textbf{0.84} \\
			\midrule
			\multirow{1}{*}{\textbf{Qwen-Max}} & \textbf{Qwen-Max} & \textbf{0.95±0.01} & \textbf{0.97±0.03} & \textbf{0.77±0.04} & \textbf{0.90} \\
			\bottomrule
		\end{tabular}%
	}
\end{table}

We investigate whether a high-capability supervisor can mitigate the limitations of smaller parameter models. We designate Qwen-Max as the universal supervisor for the heterogeneous configuration, selected for its superior performance ceiling established in Tab. \ref{tab:main_results}. Table \ref{tab:heterogeneous_results} contrasts baseline homogeneous setups against the heterogeneous configuration. The results demonstrate a universal performance uplift when sub-agents are guided by a stronger orchestrator. The most significant gain occurs in Qwen3-8B, where the average TCR nearly doubles from 0.46 to 0.70. And in sequential L2 tasks, Qwen3-8B achieves a TCR of 0.89 under supervision compared to just 0.47 autonomously. Notably, the heterogeneous combination of Qwen-Flash (sub-agent) + Qwen-Max (supervisor) achieves an average TCR of 0.84, closely approaching the SOTA performance of the standalone Qwen-Max (0.90). This improvement indicates that the primary bottleneck for smaller models is high-level task decomposition rather than atomic tool execution. By offloading complex routing logic to a strong supervisor, sub-agents can focus on concrete instruction following. Consequently, this heterogeneous approach emerges as a cost-effective strategy. It maintains high system-level accuracy through the central planner while effectively reducing computational overhead by utilizing lightweight models for execution.

\subsubsection{Efficacy of Multi-Agent Collaboration}
\begin{figure}[!t]
	\centering
	\includegraphics[width=1.0\columnwidth]{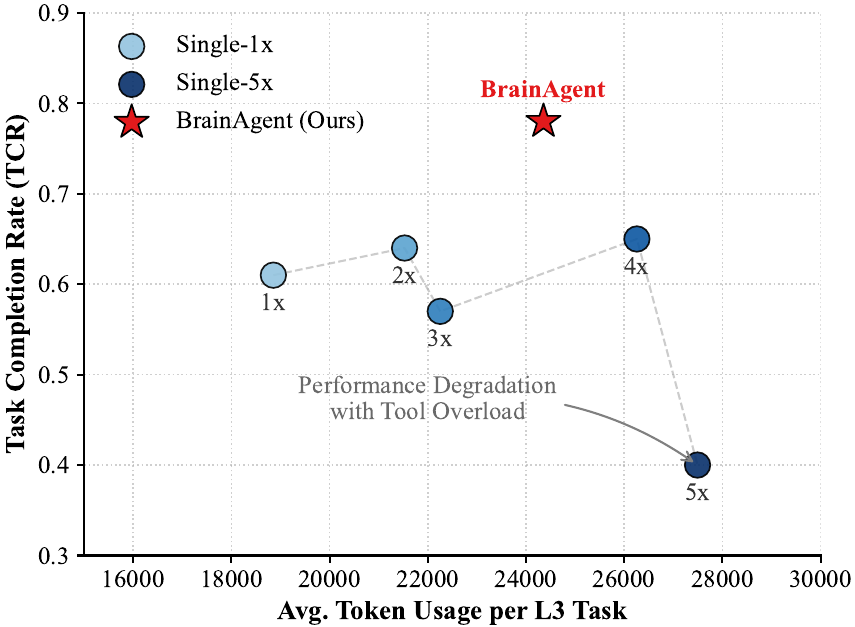}
	\caption{Efficiency--robustness trade-off on L3 tasks. BrainAgent achieves the best balance, while \texttt{SingleAgent} suffers from degraded performance and higher token consumption under increasing tool overload (1×--5×).}
	\label{fig:token_acc}
\end{figure}

To validate the efficacy of our collaborative architecture, we conducted a stress test comparing BrainAgent against a \texttt{SingleAgent} baseline, focusing on robustness and efficiency under increasing tool scales. The \texttt{SingleAgent} baseline was granted simultaneous access to the global toolset. To simulate real-world scenarios with extensive tool libraries, we introduced varying levels of distractor tools, which are functionally irrelevant but have plausible descriptions. We scaled the toolset size for the \texttt{SingleAgent} from 1× (equivalent to BrainAgent's total tool count) to 5× (original set plus 400\% distractors). In contrast, BrainAgent employs resource decoupling, ensuring sub-agents access only domain-relevant tools, thereby remaining insulated from the injected noise. Figure~\ref{fig:token_acc} illustrates the trade-off between computational cost (average tokens per task) and performance (TCR) on Level 3 tasks. BrainAgent achieves a superior TCR, significantly outperforming the unperturbed Single-1x baseline. As the toolset expands to $5\times$, the performance of the SingleAgent deteriorates drastically to 0.40, while its token consumption rises to the highest level of approximately 27.5k. This inverse correlation indicates that single agents suffer from severe attention dispersion and context overload when facing large toolsets, leading to lengthy and erroneous trial-and-error loops. Conversely, BrainAgent maintains high stability and achieves the optimal balance between high accuracy and computational efficiency, demonstrating the our design effectively mitigates the risks of tool overload.

\begin{figure}[!t]
	\centering
	\includegraphics[width=1.0\columnwidth]{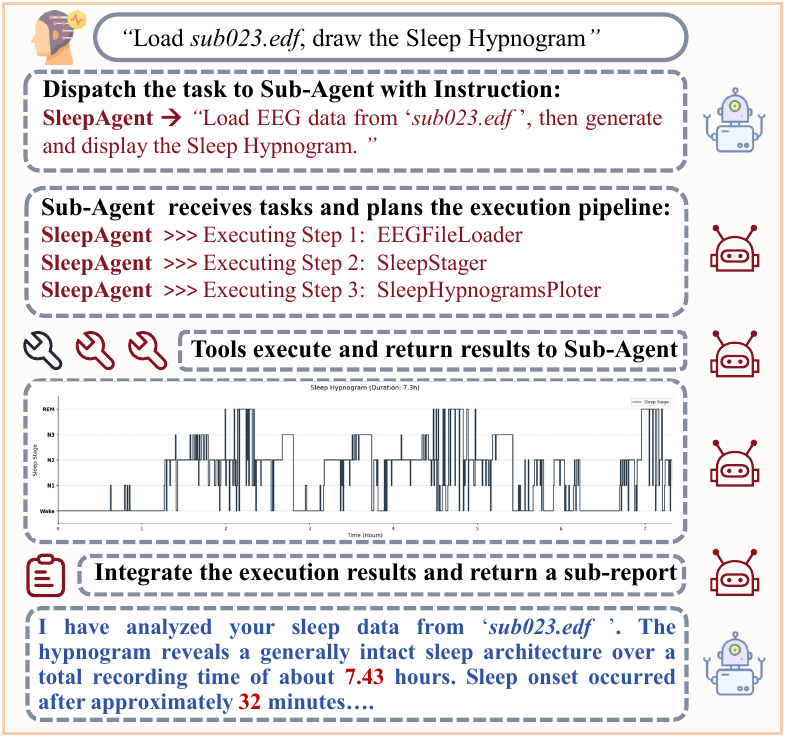}
	\caption{Visualization of an execution trace for an L1 task.}
	\label{fig:vis1}
	
\end{figure}
\begin{figure}[!t]
	\centering
	\includegraphics[width=1.0\columnwidth]{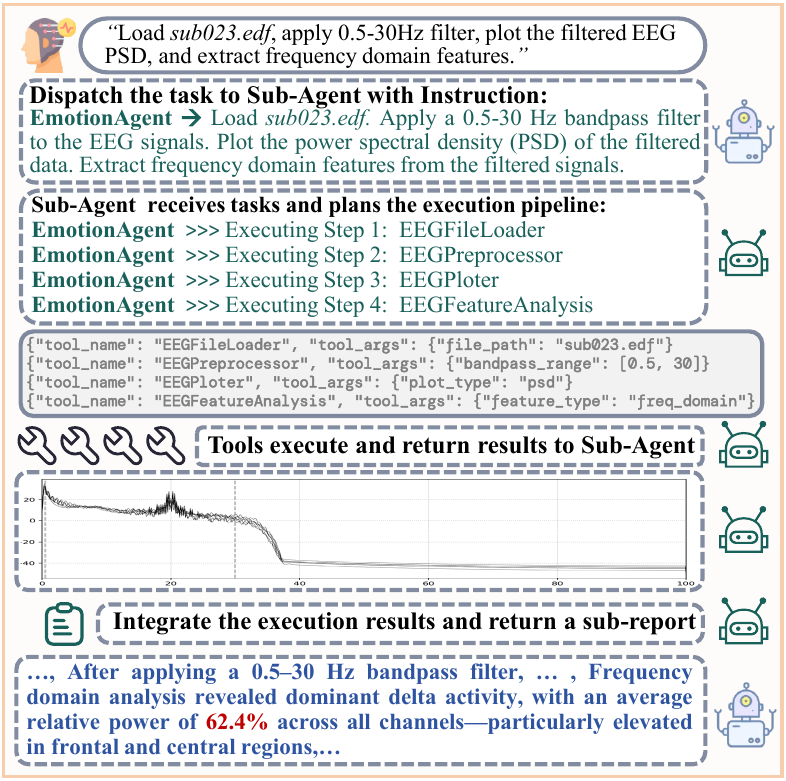}
	\caption{Visualization of an execution trace for an L2 task.}
	\label{fig:vis2}
	
\end{figure}
\begin{figure}[!t]
	\centering
	\includegraphics[width=1.0\columnwidth]{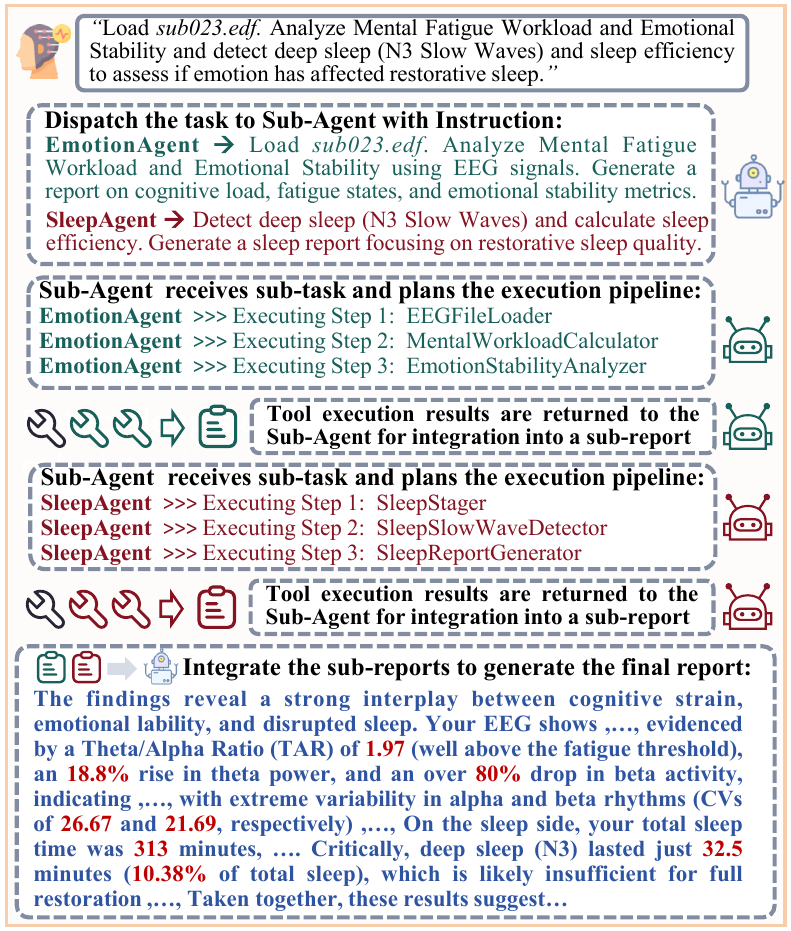}
	\caption{Visualization of an execution trace for an L3 task.}
	\label{fig:vis3}
	
\end{figure}

\subsubsection{Visualization of Execution Traces}

To qualitatively validate the decision-making process of BrainAgent, we visualize execution traces across the hierarchical difficulty spectrum (L1--L3). Figure \ref{fig:vis1} illustrates an L1 task where the user requests a sleep hypnogram. The supervisor correctly identifies the single-domain intent and dispatches the instruction to the \texttt{SleepAgent}, which autonomously constructs a linear pipeline to generate the target plot. Figure \ref{fig:vis2} expands this to an L2 scenario involving sequential signal processing with specific constraints. The \texttt{EmotionAgent} strictly adheres to user-defined parameters before executing dependent analysis steps like PSD plotting and feature extraction. Figure \ref{fig:vis3} demonstrates the system's reasoning capability in a complex hybrid task. The user queries the causal relationship between emotional states and restorative sleep quality. The supervisor decomposes this abstract human intent, assigning parallel sub-tasks: the \texttt{EmotionAgent} quantifies mental fatigue and stability, while the \texttt{SleepAgent} assesses deep sleep and efficiency. Crucially, the supervisor functions as a synthesizer rather than a mere aggregator. By integrating distinct sub-reports, it provides a comprehensive reference for further analysis.

\section{Conclusion}
We propose BrainAgent, a hierarchical multi-agent framework that translates natural language instructions into brain signal understanding. This approach lowers technical barriers, enabling clinicians and general researchers to conduct advanced analysis without specialized programming skills. Moreover, BrainAgent moves beyond static tasks by automating complex, long-horizon, and end-to-end workflows. Comprehensive benchmarking confirms that BrainAgent achieves superior reliability and robustness across complex analytical scenarios. Consequently, BrainAgent establishes a scalable baseline for intelligent BCI, democratizing access to autonomous brain signal understanding.

\bibliography{aaai2027}


\newpage
\appendix
\onecolumn
\section{Tool Descriptions} \label{tools}
\label{app:tool_descriptions}

\subsection{General-Purpose Tools}

These tools provide fundamental functionalities for data loading, preprocessing, visualization, and basic feature extraction, serving as the foundational layer accessible to all specialized agents.

\begin{itemize}
	\item \textbf{EEGFileLoader} \\
	\textit{Description:} A utility tool designed to load EEG recordings from standard file formats (e.g., .edf, .bdf) into the system memory. \\
	\textit{Arguments:}
	\begin{itemize}
		\item \texttt{file\_path} (str): The local system path to the target EEG file.
		\item \texttt{label\_path} (str): The path to the corresponding annotation file (e.g., sleep stages).
	\end{itemize}
	
	\item \textbf{EEGPreprocessor} \\
	\textit{Description:} A comprehensive signal processing tool for cleaning and preparing raw EEG data. It executes operations such as filtering, resampling, re-referencing, and artifact removal. \\
	\textit{Constraint:} Executed only when specific preprocessing steps are explicitly requested. \\
	\textit{Arguments:}
	\begin{itemize}
		\item \texttt{start\_time} (float): Start timestamp to crop the data in seconds (default: 0).
		\item \texttt{end\_time} (float): End timestamp to crop the data in seconds (default: end of recording).
		\item \texttt{resampling\_rate} (int): Target sampling rate in Hz.
		\item \texttt{bandpass\_range} (list): High-pass and low-pass frequency cutoffs (e.g., [0.5, 30]).
		\item \texttt{notch\_freq} (float): Frequency to remove line noise (e.g., 50 or 60 Hz).
		\item \texttt{reference} (str): The referencing scheme to apply (e.g., \texttt{'average'}).
		\item \texttt{channels\_to\_include} (list): A subset list of channel names to retain.
		\item \texttt{apply\_ica} (bool): Whether to perform Independent Component Analysis (ICA) for artifact removal.
		\item \texttt{ica\_components} (int): The number of ICA components to compute if ICA is enabled.
	\end{itemize}
	
	\item \textbf{EEGSaver} \\
	\textit{Description:} Saves the processed EEG data to a local file in NumPy format for external usage or persistence. \\
	\textit{Constraint:} Invoked only upon explicit user requests. \\
	\textit{Arguments:}
	\begin{itemize}
		\item \texttt{save\_path\_file} (str): Destination path including the filename.
		\item \texttt{slice\_duration} (float): Duration of each segment in seconds if segmentation is required.
	\end{itemize}
	
	\item \textbf{EEGPloter} \\
	\textit{Description:} A visualization tool for generating standard plots for signal inspection and statistical analysis. \\
	\textit{Arguments:}
	\begin{itemize}
		\item \texttt{plot\_type} (str): Specifies the visualization mode. Options include:
		\begin{itemize}
			\item \texttt{'raw'}: Time-domain waveform inspection.
			\item \texttt{'average'}: Averaged evoked potentials.
			\item \texttt{'psd'}: Power Spectral Density for frequency analysis.
			\item \texttt{'connectivity'}: Channel correlation matrix.
		\end{itemize}
	\end{itemize}
	
	\item \textbf{EEGFeatureAnalysis} \\
	\textit{Description:} A comprehensive analysis tool designed to extract multi-domain quantitative features from EEG signals and generate a clinical analysis report. \\
	\textit{Arguments:}
	\begin{itemize}
		\item \texttt{feature\_type} (str): Specifies the domain of features to extract. Options include:
		\begin{itemize}
			\item \texttt{'time\_domain'}: Statistical metrics including Mean, Std, RMS, Peak-to-Peak, Skewness, Kurtosis, Zero-Crossing Rate (ZCR), and Hjorth Parameters (Activity, Mobility, Complexity).
			\item \texttt{'freq\_domain'}: Spectral metrics including Absolute/Relative Band Power (Delta, Theta, Alpha, Beta, Gamma), Spectral Entropy, Peak Frequency, SEF95, and Cognitive Ratios (e.g., Theta/Beta).
			\item \texttt{'nonlinear'}: Complexity and chaos metrics such as Entropy and Fractal Dimension.
			\item \texttt{'connectivity'}: Functional connectivity measures (Coherence, PLV) between brain regions.
			\item \texttt{'all'} (Default): Executes all available analysis modules.
		\end{itemize}
		\item \texttt{nonlinear\_type} (str, optional): Specifies the algorithm when \texttt{feature\_type='nonlinear'}. Options:
		\begin{itemize}
			\item \texttt{'ApEn'}: Approximate Entropy (computationally intensive).
			\item \texttt{'SampEn'}: Sample Entropy (computationally intensive).
			\item \texttt{'HFD'}: Higuchi Fractal Dimension (fast computation).
			\item \texttt{'LZC'}: Lempel-Ziv Complexity (fast computation).
			\item \texttt{'all'} (Default): Calculates all nonlinear metrics.
		\end{itemize}
	\end{itemize}
	
	\item \textbf{EEGQualityAssessor} \\
	\textit{Description:} A diagnostic tool for evaluating EEG signal quality by detecting bad channels and artifacts. It outputs a quality score (0--100), a list of identified bad channels, and a binary pass/fail conclusion regarding data usability. \\
	\textit{Constraint:} Invoked only when the user explicitly requests a data quality check. \\
	\textit{Arguments:}
	\begin{itemize}
		\item \texttt{z\_threshold} (float): Z-score threshold for outlier detection (default: 3.0).
		\item \texttt{corr\_threshold} (float): Correlation threshold for detecting uncorrelated (noisy) channels (default: 0.4).
		\item \texttt{amplitude\_threshold} (float): Peak-to-peak amplitude threshold in Volts (default: $100 \times 10^{-6}$).
	\end{itemize}
\end{itemize}

\subsection*{A.2. Domain-Specific Tools (Sleep)}

These tools are exclusively managed by the \texttt{SleepAgent} and are tailored for clinical sleep analysis, including staging, microstructure detection, and hypnogram visualization.

\begin{itemize}
	\item \textbf{SleepStager} \\
	\textit{Description:} Performs automatic sleep staging inference by segmenting EEG data into fixed-length epochs and classifying them into standard stages (W, N1, N2, N3, REM). If pre-existing labels are available in the state, it prioritizes loading them over model inference. It returns the raw sequence of sleep labels. \\
	\textit{Arguments:} None.
	
	\item \textbf{SleepHypnogramsPloter} \\
	\textit{Description:} Generates and visualizes the Sleep Hypnogram (staircase plot) to depict the sleep architecture over time. \\
	\textit{Constraint:} Executed \textbf{only} if the user explicitly requests visualization. Requires prior execution of \texttt{SleepStager}. \\
	\textit{Arguments:} None.
	
	\item \textbf{SleepReportGenerator} \\
	\textit{Description:} Synthesizes a comprehensive text-based summary of the sleep architecture. Key metrics include Total Sleep Time (TST), Sleep Efficiency (SE), Sleep Latency, Wake After Sleep Onset (WASO) and etc. \\
	\textit{Constraint:} Requires prior execution of \texttt{SleepStager}. \\
	\textit{Arguments:} None.
	
	\item \textbf{SleepSpindleDetector} \\
	\textit{Description:} Automatically detects Sleep Spindles (characteristic bursts of 11--16 Hz oscillatory activity) primarily occurring during NREM sleep. \\
	\textit{Arguments:} None.
	
	\item \textbf{SleepKComplexDetector} \\
	\textit{Description:} Detects K-Complexes (high-amplitude biphasic waves) in the EEG signal. \\
	\textit{Arguments:}
	\begin{itemize}
		\item \texttt{plot\_k\_complex} (bool): Whether to visualize the detected K-Complex waveforms. Defaults to \texttt{False} unless explicitly requested by the user.
	\end{itemize}
	
	\item \textbf{SleepSlowWaveDetector} \\
	\textit{Description:} Automatically detects slow waves  characteristic of deep sleep (N3) across frontal and central channels. \\
	\textit{Constraint:} Requires prior execution of \texttt{SleepStager}. \\
	\textit{Arguments:} None.
	
	\item \textbf{SleepArousalDetector} \\
	\textit{Description:} Detects cortical arousals (abrupt shifts in EEG frequency lasting 3--15 seconds) during sleep. \\
	\textit{Constraint:} Requires prior execution of \texttt{SleepStager}. \\
	\textit{Arguments:} None.
	
	\item \textbf{SleepSpectrogramPloter} \\
	\textit{Description:} Generates a whole-night Time-Frequency Spectrogram (Hypnospectrogram) to visualize sleep depth continuity and spectral power dynamics (0.5--30 Hz). \\
	\textit{Constraint:} Executed only upon explicit visualization requests. Forbidden for text-based analysis queries. \\
	\textit{Arguments:} None.
	
	\item \textbf{SleepDisorderRiskAssessor} \\
	\textit{Description:} Performs a multi-dimensional risk assessment for common sleep disorders, including Insomnia, Apnea, Narcolepsy, Deep Sleep Deficit, and Neuro-degeneration. \\
	\textit{Constraint:} Requires results from \texttt{SleepStager} and various event detectors (Spindles, K-Complexes, Arousals) to form a complete diagnosis. \\
	\textit{Arguments:} None.
\end{itemize}

\subsection*{A.3. Domain-Specific Tools (Emotion)}

These tools are exclusively managed by the \texttt{EmotionAgent} and focus on affective computing tasks, including physiological biomarker extraction, mood stability analysis, and cognitive workload assessment and etc.

\begin{itemize}
	\item \textbf{EmotionFAAAnalyzer} \\
	\textit{Description:} Calculates Frontal Alpha Asymmetry (FAA), a key biomarker for determining emotional valence (Approach vs. Withdrawal behaviors). \\
	\textit{Constraint:} The agent must verify that the specified left and right frontal channels exist in the dataset before invocation. \\
	\textit{Arguments:}
	\begin{itemize}
		\item \texttt{left\_ch} (str): Name of the left frontal channel (e.g., \texttt{'F3'}).
		\item \texttt{right\_ch} (str): Name of the right frontal channel (e.g., \texttt{'F4'}).
	\end{itemize}
	
	\item \textbf{EmotionBiomarkerExtractor} \\
	\textit{Description:} Extracts quantitative spectral biomarkers to evaluate the user's resting-state mental health baseline. It calculates key physiological ratios including:
	\begin{itemize}
		\item Beta/Alpha Ratio (BAR): For Stress \& Arousal levels.
		\item Theta/Beta Ratio (TBR): For Attention Control \& Mind-wandering.
		\item Theta/Alpha Ratio (TAR): For Drowsiness \& Fatigue detection.
		\item Alpha Peak Frequency (APF): For Cognitive Processing Speed \& Burnout.
	\end{itemize}
	\textit{Arguments:} None.
	
	\item \textbf{EmotionStabilityAnalyzer} \\
	\textit{Description:} Evaluates the user's emotional regulation capacity and mood stability over time. It analyzes signal variability to detect mood swings and signal complexity to assess emotional rigidity, focusing on frontal and central regions. \\
	\textit{Arguments:} None.
	
	\item \textbf{MentalWorkloadCalculator} \\
	\textit{Description:} Analyzes EEG patterns associated with mental fatigue and cognitive workload. It utilizes metrics  and temporal trend analysis to detect progressive fatigue accumulation. \\
	\textit{Arguments:} None.
	
	\item \textbf{MentalHealthRiskAssessor} \\
	\textit{Description:} A rule-based synthesis tool that generates a final risk assessment for Depression, Anxiety, and Burnout (categorized as Low/Moderate/High). \\
	\textit{Constraint:} Requires the prior execution of at least two precursor tools (e.g., \texttt{EmotionFAAAnalyzer}, \texttt{EmotionBiomarkerExtractor}, \texttt{EmotionStabilityAnalyzer}, or \texttt{MentalWorkloadCalculator}) to aggregate sufficient diagnostic evidence. \\
	\textit{Arguments:} None.
\end{itemize}

\section{Knowledge Base Descriptions}
\label{app:knowledge_base}

Complementing the specialized tool mechanism, BrainAgent incorporates a hierarchical Retrieval-Augmented Generation (RAG) system. This design empowers each sub-agent with access to both a shared repository of general neurophysiological concepts and a specialized library of domain-specific protocols.

\subsection{Knowledge Decoupling Strategy}

To ensure that sub-agents generate medically accurate and contextually relevant reports, we categorize knowledge bases into two distinct layers:

\begin{itemize}
	\item \textbf{General Knowledge Base (Shared):} Accessible by all sub-agents, this repository contains fundamental concepts in Brain-Computer Interfaces (BCI) and neural signal processing. It includes definitions of standard frequency bands (e.g., Alpha, Beta), electrode placement systems (e.g., 10-20 system), and basic signal artifacts.
	\item \textbf{Domain-Specific Knowledge Base (Specialized):} Exclusive to specific sub-agents, this layer houses granular professional guidelines.
	\begin{itemize}
		\item For the \texttt{SleepAgent}, the knowledge base integrates criteria from the \textit{AASM Manual for the Scoring of Sleep and Associated Events} \cite{berry2017aasm} and clinical descriptions of sleep disorders (e.g., Apnea, Insomnia).
		\item For the \texttt{EmotionAgent}, the repository includes comprehensive reviews on affective computing, defining biomarkers such as Frontal Alpha Asymmetry (FAA) and Theta/Beta ratios for fatigue assessment.
	\end{itemize}
\end{itemize}

This hierarchical architecture supports a "plug-and-play" extension mechanism, allowing researchers to dynamically register new domain knowledge bases without altering the core system.

\subsection{Two-Stage Retrieval Mechanism}

Upon completing tool execution and preparing the final sub-report, each sub-agent queries its accessible knowledge bases to augment its findings with professional context. We implement a robust two-stage retrieval pipeline to ensure high relevance and precision:

\textbf{1. Coarse Retrieval:}

We utilize a lightweight Bi-Encoder model (\texttt{sentence-transformers/all-MiniLM-L6-v2}) to embed both the user query and knowledge documents into a shared vector space. A similarity search (e.g., FAISS) retrieves the top-N candidate documents (typically N=20) based on cosine similarity, ensuring high recall.

\textbf{2. Fine-Grained Reranking:}

To filter out irrelevant noise, we employ a Cross-Encoder model (\texttt{cross-encoder/ms-marco-MiniLM-L-6-v2}) to re-score the candidate pairs. Unlike Bi-Encoders, the Cross-Encoder processes the query and document simultaneously, capturing fine-grained semantic interactions. Candidates with scores below a calibrated threshold are discarded, and the top-K results (typically K=5) are injected into the agent's context window.

This mechanism ensures that the generated reports are not only grounded in the specific experimental results but also substantiated by authoritative medical literature, significantly reducing hallucination and enhancing clinical validity.
\section{Benchmark Dataset Examples}
\label{app:benchmark_examples}

To rigorously evaluate BrainAgent, we constructed a benchmark dataset consisting of 60 tasks stratified by difficulty (Low, Medium, High). Each entry is stored in JSON format containing the user query, input files, and the ground truth for evaluation. The complete benchmark dataset is provided in the Supplementary Material.

\subsection*{B.1. Ground Truth Schema Definition}

The \texttt{ground\_truth} field defines the success criteria for the supervisor (routing) and the sub-agents (tool execution).

\begin{itemize}
	\item \textbf{Routing Logic (\texttt{routing}):} This field is a list of lists representing the required agent dispatch.
	\begin{itemize}
		\item \textbf{OR Logic (Interchangeable):} If a sub-list contains multiple agents (e.g., \texttt{[["SleepAgent", "EmotionAgent"]]}), it implies that the task relies on \textit{Common Tools} available to both. Dispatching the task to either agent is considered correct.
		\item \textbf{AND Logic (Collaborative):} If there are multiple independent sub-lists (e.g., \texttt{[["EmotionAgent"], ["SleepAgent"]]}), the task requires cross-domain collaboration. The supervisor must dispatch sub-tasks to all listed agents to be considered correct.
	\end{itemize}
	
	\item \textbf{Execution Logic (\texttt{must\_have\_tools}):} This list defines the critical path of the execution pipeline. To achieve a successful Task Completion Rate (TCR), the generated execution trace must contain every tool listed in this set. Missing any tool from this list results in a failure.
\end{itemize}

\subsection*{B.2. Representative Cases}

\textbf{Level 1: Low Difficulty (Atomic Task)}: 
This level evaluates the agent's proficiency in executing explicit, few-step instructions using basic tools. Tasks typically involve straightforward operations such as data loading, preprocessing, or visualization. The evaluation focuses on tool parameter correctness and execution stability within a single sub-agent.
\begin{figure}[!h]
	\centering
	\includegraphics[width=0.93\columnwidth]{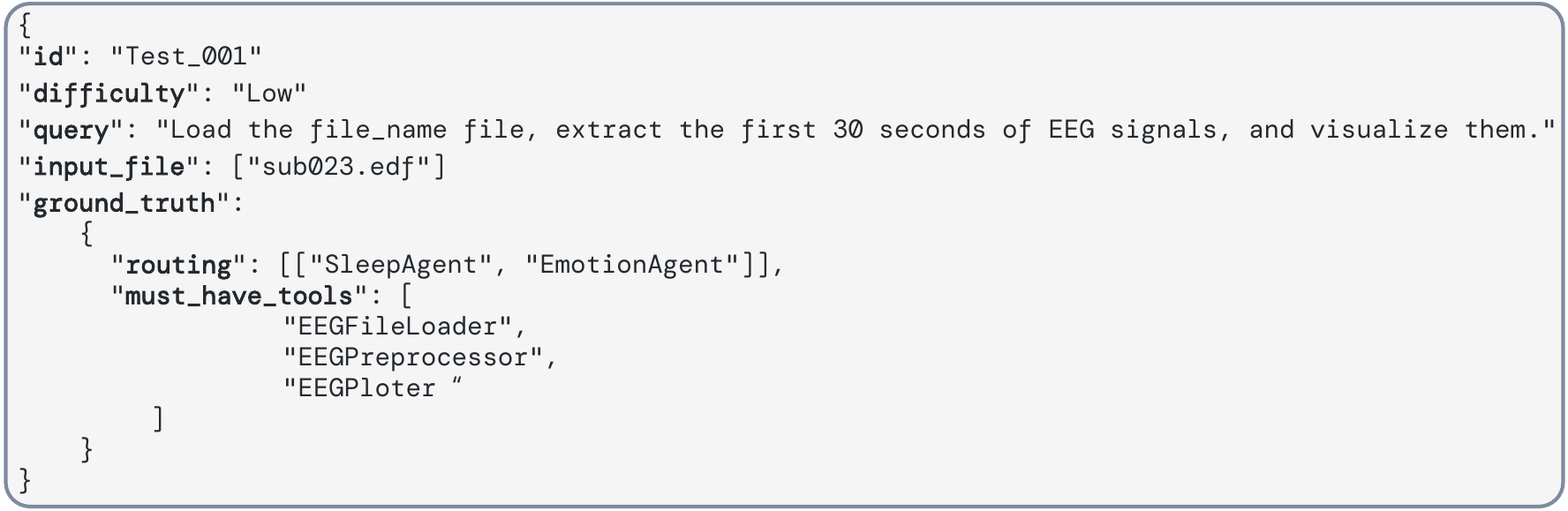}
	\caption{A representative case of an L1 task.}
	\label{fig:case1}
\end{figure}

\textbf{Level 2: Medium Difficulty (Sequential Task)}
This level assesses the agent's ability to plan and execute long-horizon workflows within a single domain, testing its capacity to accurately interpret and execute complex, multi-step instructions. These tasks imply strict logical dependencies where the output of one tool serves as the prerequisite for another. Success requires the agent to adhere to a precise operational sequence, ensuring no critical tools are omitted while avoiding the hallucination of non-existent dependencies.

\begin{figure}[!h]
	\centering
	\includegraphics[width=0.93\columnwidth]{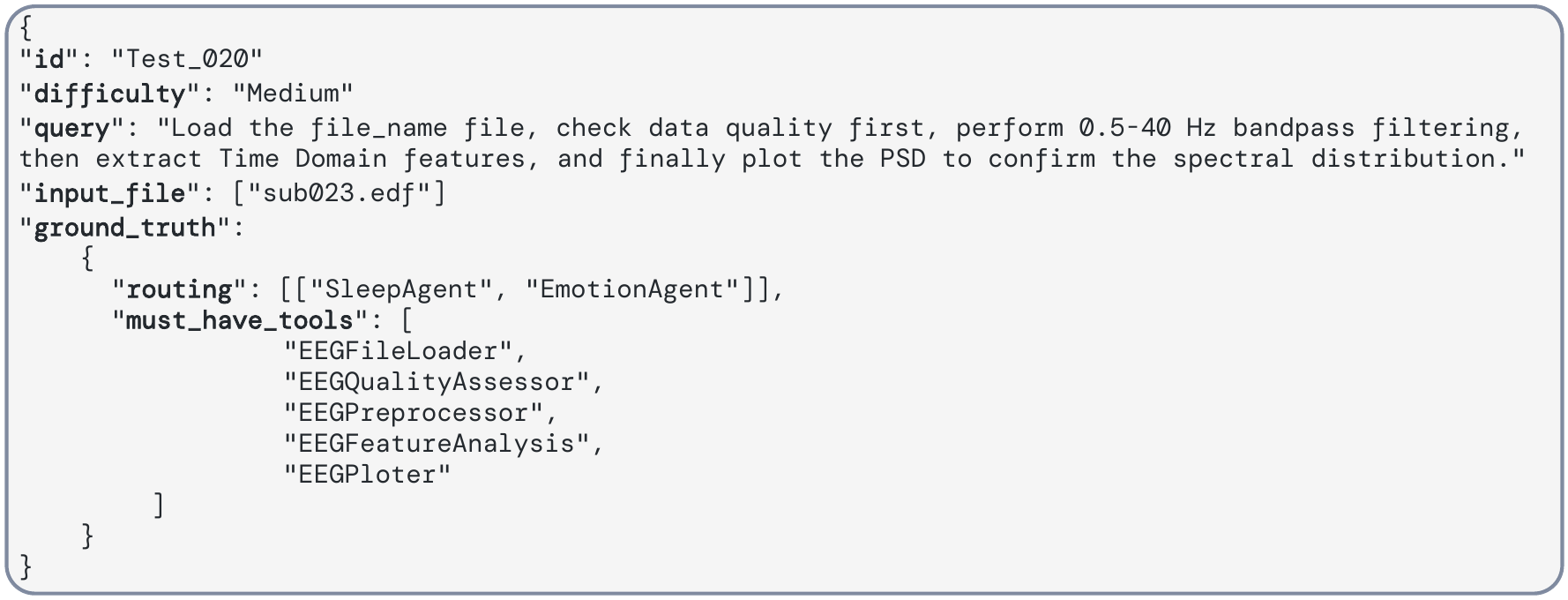}
	\caption{A representative case of an L2 task.}
	\label{fig:case2}
\end{figure}

\textbf{Level 3: High Difficulty (Hybrid Task)}:
This level represents the highest complexity, simulating real-world scenarios driven by ambiguous, high-level user intents rather than explicit instructions. Tasks often involve abstract queries that necessitate complex, long-horizon workflows and cross-domain collaboration. Success depends on the system's ability to decompose abstract goals, accurately coordinate multiple agents, and synthesize heterogeneous outputs into a coherent analysis.

\begin{figure}[!h]
	\centering
	\includegraphics[width=0.93\columnwidth]{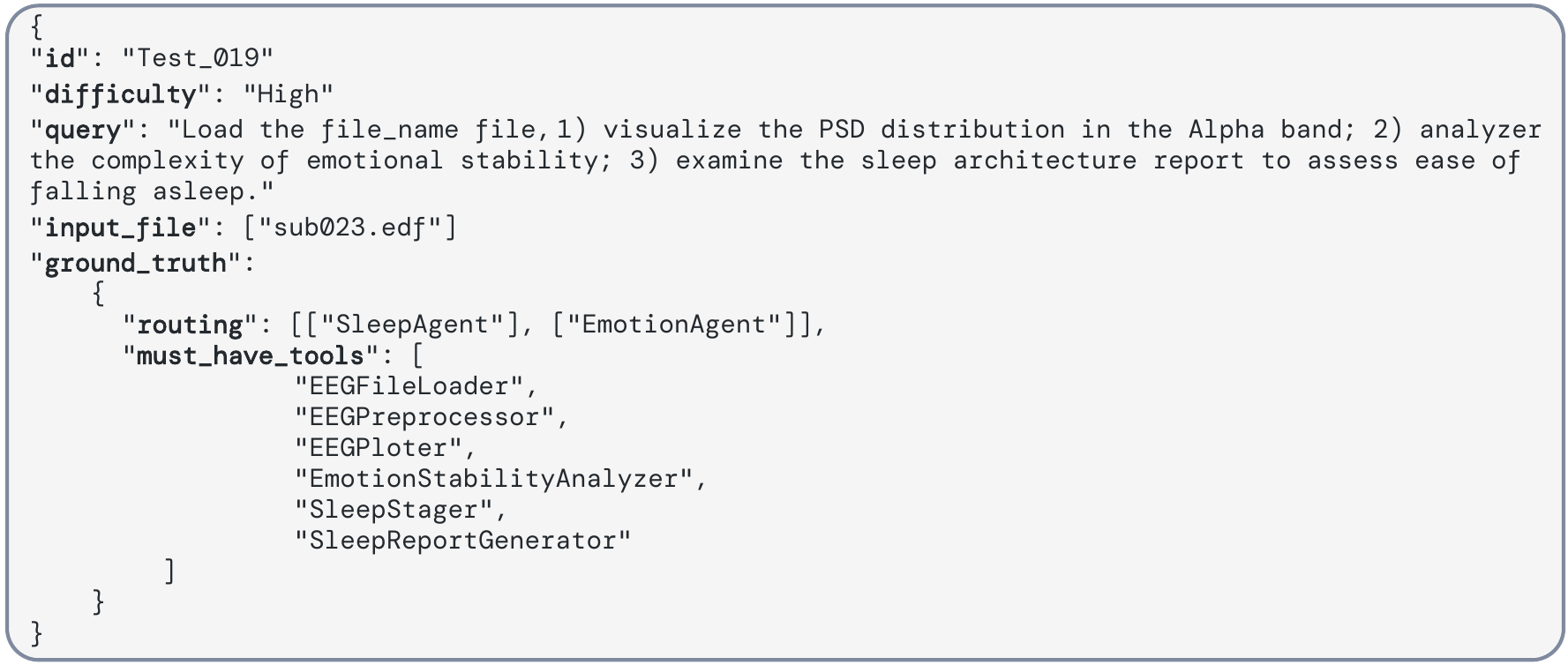}
	\caption{A representative case of an L3 task.}
	\label{fig:case3}
\end{figure}

\section{Agent System Prompts}
\label{app:prompts}

To ensure reproducibility and transparency, we provide the core system prompts designed for the Supervisor and Sub-agents. Note that dynamic contents (e.g., specific tool lists, conversation history) are represented by placeholders (e.g., \texttt{\{...\}}).

\subsection{Supervisor Agent Dispatch Prompt}
The supervisor agent acts as the central orchestrator. It uses the following system prompt to enforce strict routing strategies, distinguishing between atomic operations and composite analysis pipelines while maintaining a stateless routing logic.

\begin{tcolorbox}[colback=gray!5!white,colframe=gray!50!black,title=\textbf{System Prompt: Supervisor Agent Dispatching},fonttitle=\bfseries,arc=0mm,boxrule=0.5pt]
	\small
	\textbf{\#\#\# ROLE DEFINITION} \\
	You are the \textbf{Central Orchestrator (Supervisor)} for an EEG Analysis System. Your goal is to parse user requests and dispatch actionable tasks to the appropriate Sub-Agents.
	
	\textbf{\#\#\# AVAILABLE SUB-AGENTS} \\
	\texttt{\{agent\_list\}}
	
	\textbf{\#\#\# CONTEXT DATA} \\
	\textbf{History:} \texttt{\{conversation\_history\}} \\
	\textbf{Shared State:} \texttt{\{current\_system\_state\}}
	
	\vspace{0.5em}
	\textbf{Analyze the User Request:} "\texttt{\{user\_input\}}" and decide:
	\begin{enumerate}[nosep,leftmargin=*]
		\item What does the user want to do?
		\item Which Agent needs to be called?
		\item If no Agent call is needed (e.g., greetings, capability inquiries), respond directly.
	\end{enumerate}
	
	\vspace{0.5em}
	\textbf{\#\#\# ROUTING STRATEGY}
	\begin{itemize}[leftmargin=*,itemsep=2pt]
		\item \textbf{1. Single Domain Mapping (Direct Assignment):} The user's intent focuses exclusively on one specific aspect of their physiological state.
		
		\item \textbf{2. Multiple Domain Mapping (Sequential Orchestration):} The request involves multiple aspects or implies a causal relationship. Schedule tasks for ALL relevant agents and determine the logical execution order (e.g., Causal Analysis: Cause $\rightarrow$ Effect; Exploratory: General $\rightarrow$ Specific).
		
		\item \textbf{3. Atomic vs. Composite Operations (CRITICAL):}
		\begin{itemize}[nosep]
			\item \textbf{Atomic Operations (Strict Execution):} If the user requests a specific, single-step action (e.g., "Load the file"), you must generate a task for \textbf{THAT ACTION ONLY}. Do NOT assume or append downstream steps like "filtering" or "analysis" unless explicitly requested.
			\item \textbf{Composite/Full Pipeline:} Only if the user asks for a high-level goal (e.g., "Analyze this data") should you include preprocessing and analysis steps.
		\end{itemize}
		
		\item \textbf{4. Direct Response Protocol (No Dispatch Needed):} Set \texttt{needs\_dispatch} to \texttt{False} for Chitchat, Capability Inquiries, or Information Retrieval (if the answer is already in History/Shared State).
	\end{itemize}
	
	\vspace{0.5em}
	\textbf{\#\#\# STRICT CONSTRAINTS}
	\begin{enumerate}[leftmargin=*,itemsep=2pt]
		\item \textbf{Atomic Execution Principle:} You must strictly adhere to the \textbf{Explicit Intent} of the user's directive. \textbf{DO NOT} perform "helpful" look-ahead actions.
		\item \textbf{Passive Reactivity:} You are a router, not a planner. Do not attempt to complete a full analysis pipeline in one turn unless explicitly requested. If a step requires prerequisites not done, simply issue the prerequisite task.
		\item The \texttt{task\_content} must be a self-contained command that the Sub-Agent can execute blindly.
	\end{enumerate}
	
	\vspace{0.5em}
	\textbf{\#\#\# OUTPUT FORMAT} \\
	Respond with a single JSON object. Do not use Markdown formatting.
	\begin{verbatim}
		{
			"thought_process": "Brief analysis of user intent, explicitly stating 
			if the request is Atomic or Composite.",
			"needs_dispatch": true, 
			"direct_response": null,
			"task_queue": [
			{
				"target_agent": "sleep_agent", 
				"task_content": "Strictly constrained instruction."
			}
			]
		}
	\end{verbatim}
\end{tcolorbox}

\subsection{Sub-Agent Execution Prompts}
All specialized sub-agents (e.g., SleepAgent, EmotionAgent) share a unified execution logic defined by the following prompts. The specific capabilities of each agent are dynamically injected via the \texttt{\{available\_tools\_str\}} placeholder.

\begin{tcolorbox}[colback=gray!5!white,colframe=gray!50!black,title=\textbf{System Prompt: Sub-Agent Executing},fonttitle=\bfseries,arc=0mm,boxrule=0.5pt]
	\small
	\textbf{\#\#\# ROLE DEFINITION} \\
	\texttt{\{system\_role\}} \\
	\textbf{\#\#\# TOOL USAGE PROTOCOL} \\
	You have access to the following tools: \texttt{\{available\_tools\_str\}} \\
	All tools operate on a global \texttt{shared\_state} dictionary. They read inputs from it and write outputs back to it.
	
	\textbf{\#\#\# RESPONSIBILITIES}
	\begin{enumerate}[nosep,leftmargin=*]
		\item Analyze the task requirements.
		\item Create a reasonable execution plan.
		\item Use appropriate tools to complete the task.
		\item Provide professional analysis and interpretation.
	\end{enumerate}
	
	\vspace{0.5em}
	\textbf{\#\#\# 1. ASSIGNED TASK} \\
	Directive: \texttt{\{task\_description\}}
	
	\textbf{\#\#\# 2. WORKFLOW CONTEXT} \\
	State data produced by previous sub-agents:  \texttt{\{shared\_context\}}
	
	\textbf{\#\#\# 3. CURRENT SYSTEM STATE} \\
	\texttt{\{current\_info\_str\}}
	
	\vspace{0.5em}
	\textbf{\#\#\# 4. EXECUTION DECISION PROTOCOL (CRITICAL)} \\
	\textbf{Constraint:} Strictly focus on the 'ASSIGNED TASK'. \textbf{Do NOT} execute any tasks not explicitly requested.
	
	\textbf{STEP A: Quick Check — Can You Answer Directly? (ALWAYS DO THIS FIRST)} \\
	Before any planning, ask yourself: Can I fully answer this task using ONLY the information in \texttt{WORKFLOW CONTEXT} or \texttt{CURRENT SYSTEM STATE}?
	\begin{itemize}[nosep,leftmargin=*]
		\item \textbf{If YES $\rightarrow$} Stop here. Set \texttt{"plan": []} and provide the answer in \texttt{"thought"}. Skip remaining steps.
		\item \textbf{If NO $\rightarrow$} Proceed to STEP B.
	\end{itemize}
	
	\textbf{STEP B: Task Decomposition \& Full Planning (Only if action required)} \\
	Since tool execution is needed, you MUST:
	\begin{enumerate}[nosep,leftmargin=*]
		\item \textbf{Decompose:} Break down the Directive into atomic sub-goals.
		\item \textbf{Identify:} List ALL tools required for each sub-goal.
		\item \textbf{Order:} Determine execution sequence based on data dependencies.
		\item \textbf{Plan Completely:} Generate the FULL tool sequence in ONE response.
	\end{enumerate}
	\textbf{CRITICAL:} Do NOT plan only the first step. Plan and output EVERY step from start to finish.
	
	\textbf{STEP C: Parameter Rules}
	\begin{enumerate}[nosep,leftmargin=*]
		\item \textbf{User-Specified:} If directive specifies params (e.g., "filter 1-30Hz") $\rightarrow$ include in \texttt{tool\_args}.
		\item \textbf{Defaults:} If not specified $\rightarrow$ use \texttt{tool\_args: \{\}}.
		\item \textbf{Key Consistency:} Use exact key names from \texttt{shared\_state}.
	\end{enumerate}
	\textbf{Path Safety:} Validate file paths before loading.
	
	\vspace{0.5em}
	\textbf{\#\#\# 5. OUTPUT FORMAT (STRICT JSON)} \\
	Respond with a SINGLE JSON object. Do not include markdown blocks.
	
	\textbf{Example 1: No Action Needed (Shortcut)}
	\begin{verbatim}
		{
			"thought": "Checking CURRENT SYSTEM STATE: sfreq = 256 Hz. 
			Answer available directly. No tool needed.",
			"plan": []
		}
	\end{verbatim}
	
	\textbf{Example 2: Multi-Step Complex Task}
	\begin{verbatim}
		{
			"thought": "Checked state: no EEG loaded. Steps: 1) Load -> 
			2) Sleep staging -> 3) Report. Planning all steps now.",
			"plan": [
			{ "tool_name": "EEGFileLoader", "tool_args": { "file_path": "..." } },
			{ "tool_name": "SleepStager", "tool_args": {} },
			{ "tool_name": "SleepReportGenerator", "tool_args": {} }
			]
		}
	\end{verbatim}
\end{tcolorbox}

\subsection{Sub-Agent Report Generation Prompt}
After executing the tool pipeline, the sub-agent utilizes the following prompt to synthesize raw execution outputs into a structured report for the supervisor.

\begin{tcolorbox}[colback=gray!5!white,colframe=gray!50!black,title=\textbf{System Prompt: Sub-Agent Reporting},fonttitle=\bfseries,arc=0mm,boxrule=0.5pt]
	\small
	\textbf{\#\#\# ROLE DEFINITION} \\
	\texttt{\{system\_role\}} \\
	Your current role is to synthesize execution results into a final report for the Supervisor.
	
	\textbf{\#\#\# RESPONSIBILITIES}
	\begin{enumerate}[nosep,leftmargin=*]
		\item Analyze the original task requirements.
		\item Synthesize technical tool outputs into functional summaries.
		\item Provide professional analysis and interpretation based on reference knowledge.
		\item Ensure the report is comprehensive enough for the Supervisor to answer the user without follow-up queries.
	\end{enumerate}
	
	\vspace{0.5em}
	\textbf{[SUB-AGENT FINAL REPORT GENERATION]}
	
	\textbf{\#\#\# CONTEXT \& INPUTS}
	
	\textbf{1. Assigned Task:} \\
	\texttt{\{task\_description\}} 
	\textit{(The specific directive you just executed)}
	
	\textbf{2. Workflow Context (Shared State):} \\
	\texttt{\{shared\_context\}} 
	\textit{(State data produced by previous steps/peers)}
	
	\textbf{3. Execution Outputs:} \\
	\texttt{\{execution\_results\}} 
	\textit{(Technical results from your tools)}
	
	\textbf{4. Reference Knowledge:} \\
	\texttt{\{knowledge\_str\}} 
	\textit{(Background rules or contextual information for interpretation)}
	
	\textbf{5. Current System State:} \\
	\texttt{\{current\_info\_str\}} 
	\textit{(Currently loaded data and configuration parameters)}
	
	\vspace{0.5em}
	\textbf{\#\#\# CONTENT GUIDELINES}
	\begin{enumerate}[nosep,leftmargin=*]
		\item \textbf{Abstraction:} Convert low-level technical details into functional summaries.
		\item \textbf{Clarity:} Avoid raw dictionaries. Use plain, high-level language suitable for quick review.
		\item \textbf{Focus:} Highlight the \textit{outcome} (Success/Failure) and the \textit{significance} of the result.
	\end{enumerate}
	
	\vspace{0.5em}
	\textbf{\#\#\# OUTPUT FORMAT (STRICT JSON)} \\
	Strictly output a single JSON object. Do not include markdown blocks.
	
	\begin{verbatim}
		{
			"status": "success" | "error",
			"agent": "Agent Name",
			"task": "Brief repetition of the task",
			"result": "A comprehensive and detailed report. Include specific values, 
			key observations, and clinical interpretations based on Reference Knowledge."
		}
	\end{verbatim}
\end{tcolorbox}

\subsection{Sub-Agent Report Generation Prompt}
\begin{figure}[!ht]
	\centering
	\includegraphics[width=0.5\columnwidth]{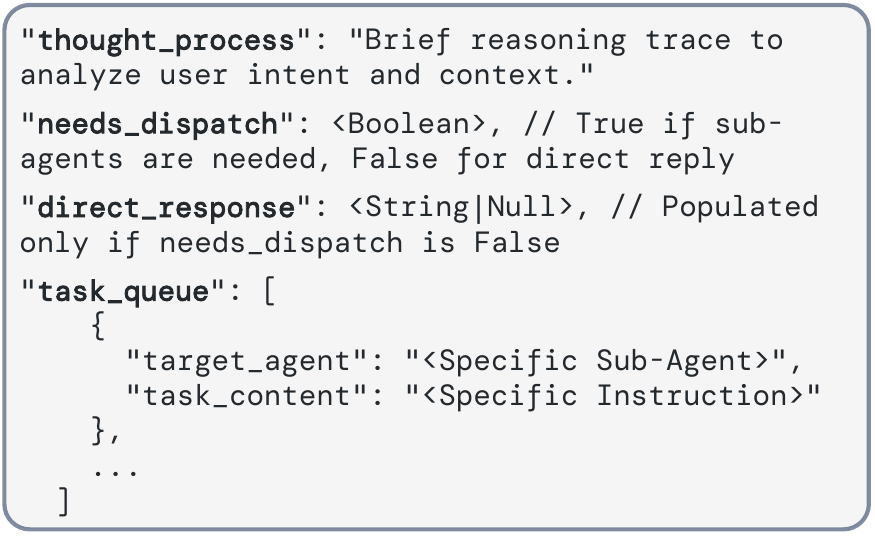}
	\caption{The JSON schema for the supervisor's decision output. The \texttt{needs\_dispatch} flag acts as a gatekeeper, determining whether to generate a \texttt{direct\_response} or populate the \texttt{task\_queue} for delegation.}
	\label{fig:supervisor}
\end{figure}

\begin{figure}[!ht]
	\centering
	\includegraphics[width=0.5\columnwidth]{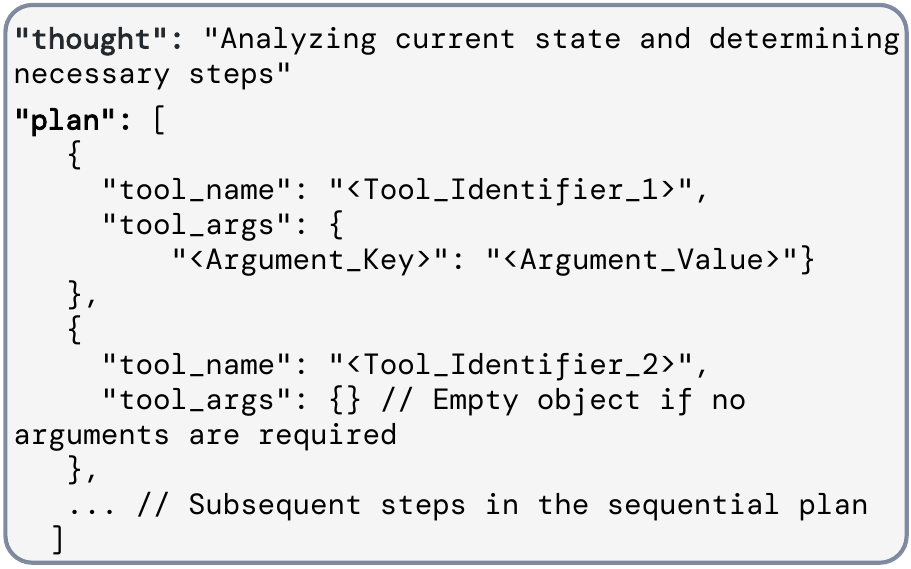}
	\caption{The JSON schema for the sub-agent's decision output. The \texttt{thought} field captures the reasoning trace, while the \texttt{plan} list contains a sequence of executable tool calls defined by their names and arguments.}
	\label{fig:sub-agent}
\end{figure}

The JSON schema of the supervisor’s and sub-agent's decision outputs are as illustrated in Fig.~\ref{fig:supervisor}, \ref{fig:sub-agent}.

\section{Output Quality Evaluation}

In addition to task completion, we further evaluate whether BrainAgent produces factually correct outputs. Since there is currently no standard benchmark for assessing agent-generated clinical narratives in agentic BCI systems, we construct a targeted output-quality evaluation on ISRUC Subgroup-1. Specifically, we select five subjects (IDs 001--005) and design nine test cases covering numerical feature extraction, spectral analysis, channel selection, ranked sequence prediction, categorical decision making, and file-level consistency. For each case, we independently implement a reference pipeline to obtain the ground-truth result. These reference pipelines are manually verified and are independent of the BrainAgent execution workflow. BrainAgent is then asked to complete the same task using Qwen3-Max, and the key outputs are extracted from its final response or generated files for comparison.

\begin{table*}[ht]
	\centering
	\caption{Overview of the output quality evaluation cases.}
	\label{tab:output_quality_cases}
	\begin{tabular}{l p{0.9\textwidth}}
		\toprule
		Case & Task Description \\
		\midrule
		Case 1 & Extract EEG channels, apply a 0.5--40 Hz bandpass filter, and compute the average alpha-band relative power across EEG channels. \\
		Case 2 & Select frontal EEG channels, apply a 1--40 Hz bandpass filter, and compute the Alpha/Beta Ratio (ABR) and Theta/Beta Ratio (TBR). \\
		Case 3 & Select central EEG channels, apply a 50 Hz notch filter and a 0.5--45 Hz bandpass filter, resample to 100 Hz, and compute Hjorth Mobility and Hjorth Complexity using the first 90 seconds. \\
		Case 4 & Select central and occipital EEG channels, apply a 0.5--40 Hz bandpass filter and a 60 Hz notch filter, resample to 200 Hz, and compute SEF95 for each region. \\
		Case 5 & Compute whole-night sleep efficiency (SE). \\
		Case 6 & Select EEG channels, apply a 0.5--40 Hz bandpass filter, compute pairwise Pearson correlations, and return the top three channel pairs ranked from highest to lowest. \\
		Case 7 & Identify the dominant and secondary EEG frequency bands in the 110--120 min segment. \\
		Case 8 & Select EEG channels, identify the channel with the highest alpha relative power, apply an 8--13 Hz filter, and save the processed signal as \texttt{{subject}\_process.npy}. \\
		Case 9 & Select EEG channels, use the first 30 seconds, apply an 8--13 Hz filter, compute Pearson correlations, and report the global synchronization value and the top correlated channel pair. \\
		\bottomrule
	\end{tabular}
\end{table*}

\begin{table*}[t]
	\centering
	\caption{Output quality evaluation results. Each cell reports ground truth / BrainAgent output.}
	\label{tab:output_quality_results}
	\resizebox{\textwidth}{!}{
		\begin{tabular}{l l c c c c c}
			\toprule
			Case & Feature & S1 & S2 & S3 & S4 & S5 \\
			\midrule
			Case 1 & Alpha ratio (\%) & 11.82 / 11.97 & 8.72 / 8.39 & 11.93 / 11.60 & 5.88 / 5.70 & 7.41 / 7.61 \\
			Case 2 & ABR & 2.29 / 2.43 & 0.98 / 1.01 & 1.18 / 1.21 & 1.19 / 1.22 & 0.68 / 0.71 \\
			Case 2 & TBR & 3.72 / 3.94 & 0.99 / 1.03 & 1.68 / 1.75 & 2.07 / 2.13 & 0.76 / 0.81 \\
			Case 3 & Hjorth Mobility & 0.58 / 0.57 & 0.76 / 0.76 & 0.68 / 0.68 & 0.60 / 0.60 & 0.75 / 0.75 \\
			Case 3 & Hjorth Complexity & 2.33 / 2.35 & 1.81 / 1.81 & 1.76 / 1.76 & 2.37 / 2.36 & 1.96 / 1.96 \\
			Case 4 & SEF95-central & 20.88 / 20.75 & 18.25 / 18.13 & 18.63 / 18.50 & 13.75 / 13.88 & 21.50 / 21.38 \\
			Case 4 & SEF95-occipital & 14.00 / 13.75 & 16.12 / 16.00 & 17.38 / 17.75 & 17.25 / 17.13 & 25.25 / 25.25 \\
			Case 5 & SE (\%) & 70.00 / 62.84 & 74.59 / 59.79 & 86.00 / 75.26 & 97.09 / 91.50 & 66.17 / 57.33 \\
			Case 6 & Top-3 pairs &
			[C4-F4,F3-F4,C4-O2] / [C4-F4,F3-F4,C4-O2] &
			[C4-F4,C3-F3,F3-F4] / [C4-F4,C3-F3,F3-F4] &
			[C3-F3,C4-F4,F3-F4] / [C3-F3,C4-F4,O2-O1] &
			[C3-F3,C4-F4,F3-F4] / [C3-F3,C4-F4,C4-C3] &
			[C3-F3,F4-C4,C4-O2] / [F3-C3,F4-C4,LOCA2-F3] \\
			Case 7 & Dominant band & delta / delta & delta / delta & delta / delta & delta / delta & delta / delta \\
			Case 7 & Secondary band & theta / theta & theta / theta & alpha / theta & theta / theta & beta / beta \\
			Case 8 & File status & Pass & Pass & Pass & Fail & Pass \\
			Case 9 & Global sync & 0.3061 / 0.3068 & 0.4315 / 0.4314 & 0.3047 / 0.3336 & 0.2913 / 0.2438 & 0.3542 / 0.3408 \\
			Case 9 & Top pair & C4-F4 / C4-F4 & C4-F4 / C4-F4 & C4-F4 / C4-F4 & C3-F3 / C3-F3 & C3-F3 / C3-F3 \\
			\bottomrule
		\end{tabular}
	}
\end{table*}

As shown in Table~\ref{tab:output_quality_results}, BrainAgent produces highly consistent results for most numerical signal-processing features, including alpha relative power, ABR/TBR, Hjorth parameters, SEF95, and global synchronization. It also correctly identifies the dominant frequency band for all subjects and the top correlated channel pair in Case 9 for all subjects. The main errors occur in sleep efficiency estimation, fine-grained top-3 channel-pair ranking, and one file-generation case, suggesting that annotation-dependent metrics, ranked sequence outputs, and artifact generation remain more challenging than direct signal-feature computation. Overall, this evaluation verifies the factual correctness of BrainAgent outputs from multiple perspectives, including numerical fidelity, discrete decision correctness, sequence-level consistency, and file-level validity.

\section{Error Analysis and Case Studies}
\label{app:error_analysis}

To provide a transparent evaluation of BrainAgent's limitations and facilitate future research, we conduct a qualitative analysis of failure cases. We categorize errors into two distinct archetypes: capability deficits inherent to lightweight models and reasoning anomalies observed in frontier models under high complexity.

\subsection{Capability Deficits in Lightweight Models}
Despite the structural support of the BrainAgent framework, small-scale models struggle with strict schema adherence and long-horizon dependencies. In this subsection, we identify three primary failure modes as follows:

\textbf{JSON Syntax Fragility.}
For models that lack native function-calling fine-tuning, generating strictly structured JSON output poses a significant challenge. Common failures include:
\begin{itemize}
	\item \textbf{Markdown Pollution:} Wrapping the JSON in markdown code blocks (e.g., \texttt{```json ... ```}) despite explicit prohibitions in the system prompt.
	\item \textbf{Syntax Errors:} Missing closing brackets, unescaped quotes within strings, or trailing commas, which cause the downstream parser to fail.
\end{itemize}

\textbf{Tool Hallucination.}
In complex Level 3 tasks, small models often fail to decompose a high-level intent into a sequence of existing atomic tools. Instead, they hallucinate non-existent "super-tools" that conveniently solve the entire problem in one step.
\begin{itemize}
	\item \textbf{Case:} User asks for a comprehensive sleep quality assessment.
	\item \textbf{Failure:} The agent attempts to call \texttt{SleepQualityAssessor} or \texttt{SleepFeatureExtractor}.
	\item \textbf{Analysis:} These tools do not exist in the provided toolset. The model hallucinates them based on semantic similarity to the user's query, revealing a failure in grounding, which is the ability to map intent strictly to the available action space.
\end{itemize}

\textbf{Parameter Hallucination.}
Even when the correct tool is selected, small models frequently struggle to adhere to the specific argument schema defined in the context, often reverting to generic coding conventions from their pre-training data.
\begin{itemize}
	\item \textbf{Case:} User requests a bandpass filter (0.5--30 Hz).
	\item \textbf{Ground Truth:} \texttt{"tool\_args": \{"bandpass\_range": [0.5, 30]\}}
	\item \textbf{Failure:}
	\begin{verbatim}
		{
			"tool_name": "EEGPreprocessor",
			"tool_args": {
				"l_freq": 0.5,   // Hallucinated key 
				"h_freq": 30     // Hallucinated key
			}
		}
	\end{verbatim}
	\item \textbf{Analysis:} The model ignores the in-context definition (\texttt{bandpass\_range}) and hallucinates parameters (e.g., \texttt{l\_freq}) likely learned from open-source libraries during pre-training. This indicates that the model's parametric knowledge overpowers its in-context learning capability.
\end{itemize}

\subsection{Reasoning Anomalies in Frontier Models}
While frontier models demonstrate superior adherence to syntactic schemas and instruction following, they exhibit subtle logical failures when handling domain-specific workflows. Unlike lightweight models that struggle with structure, these models fail at inferring unstated prerequisites.

\textbf{Implicit Dependency Neglect.}
In brain signal analysis, certain operations strictly require intermediate data transformation steps that are rarely explicitly stated in natural language instructions. Large models often map user verbs directly to tools, skipping these latent physical prerequisites. A representative instance is the Level 3 task defined in Figure \ref{fig:case3}.

\begin{figure}[!t]
	\centering
	\includegraphics[width=1.0\columnwidth]{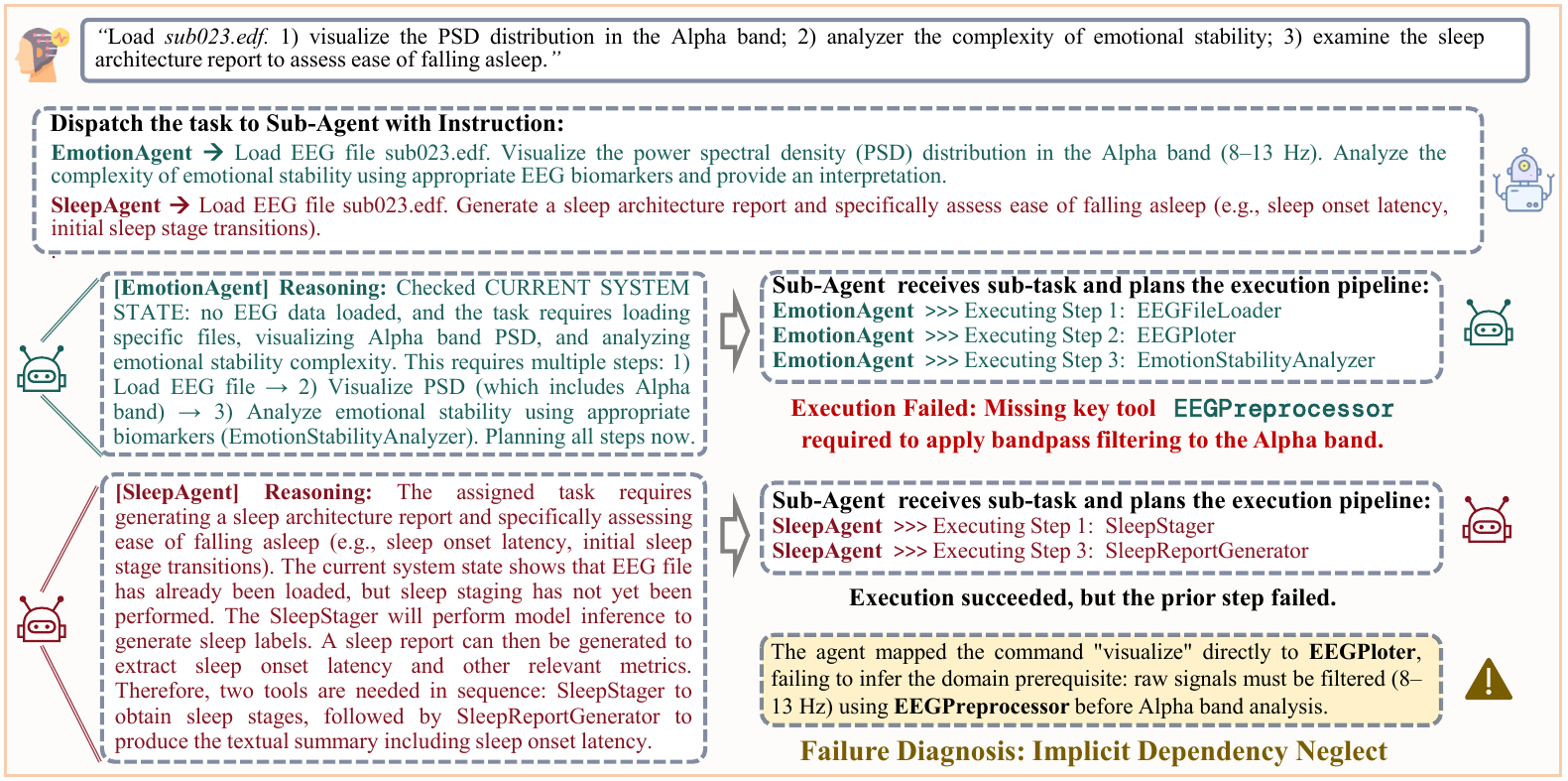}
	\caption{A failure case analysis of \textbf{Qwen3-Max} on a Level 3 task. Although the agent correctly planned the high-level steps, it failed to execute the implicit preprocessing step (Bandpass Filter) required for Alpha band visualization, illustrating the challenge of \textbf{Implicit Dependency Neglect}.}
	\label{fig:fail_case}
\end{figure}

\begin{itemize}
	\item \textbf{Case:} The detailed execution trace is shown in Figure \ref{fig:fail_case}, the user instructs the agent to visualize the PSD distribution in the Alpha band.
	\item \textbf{Execution Plan:} The \texttt{EmotionAgent} decomposes this into:
	\begin{enumerate}
		\item \texttt{EEGFileLoader} (Load data)
		\item \texttt{EEGPlotter} (Visualize data)
		\item \texttt{EmotionStabilityAnalyzer} (Analyze complexity)
	\end{enumerate}
	\item \textbf{Failure:} The execution fails because the pipeline lacks an \texttt{EEGPreprocessor} step. Raw EEG signals contain broadband frequencies; isolating the Alpha band (8--13 Hz) necessitates a bandpass filter before any visualization or spectral analysis can occur.
	\item \textbf{Analysis:} The error stems from the model's inability to bridge the gap between semantic intent and domain logic. The user input contains an \textbf{implicit} requirement to analyze the ‘Alpha band’, but does not \textbf{explicitly} instruct the agent to ‘perform 8--13 Hz bandpass filtering’ prior to visualization. Consequently, the model exhibits \textbf{Direct Mapping Bias}: it links the verb ‘visualize’ directly to the plotting tool, neglecting the domain-specific necessity of isolating the frequency band via \texttt{EEGPreprocessor} before spectral analysis.
\end{itemize}
\section{Discussion}
\label{app:discussion}

In this section, we discuss the implications of our findings for the neurotechnology community and examine the limitations that guide future research.

\subsection{Implications}
BrainAgent significantly advances the democratization of brain signal analysis. By enabling the translation of natural language into complex, long-horizon, and end-to-end executable pipelines, the framework empowers clinicians and researchers to perform rigorous investigation without the constraints of technical programming barriers. Furthermore, the collaboration between a capable supervisor and specialized sub-agents ensures that expensive reasoning resources are allocated efficiently. Rather than relying on rigid monolithic models, this modular, collaborative design enables the framework to handle increasingly complex analytical scenarios with superior efficiency. Crucially, such high extensibility ensures that BrainAgent can seamlessly adapt to evolving analytical requirements and scale across diverse brain signal modalities.

\subsection{Limitations}
Despite its capabilities, BrainAgent currently has two primary limitations that outline our future roadmap. First is the operational mode. The current framework performs retrospective analysis in an offline setting. Future work will focus on evolving this architecture into a real-time, online agentic system to support instantaneous BCI applications. Second is the scope of signal modalities. While effective for current tasks, we aim to broaden the framework's applicability by integrating analytical tools for diverse brain signal modalities, such as fMRI and fNIRS, thereby achieving a more universal platform for brain signal understanding.
\end{document}